\documentclass[11pt,a4paper]{article}
\usepackage[T1]{fontenc}
\usepackage[utf8]{inputenc}
\usepackage{newtxtext,newtxmath}
\usepackage[scaled=0.94]{helvet}
\usepackage[left=23mm,right=23mm,top=29mm,bottom=25mm,headheight=36pt,headsep=12pt,footskip=28pt]{geometry}
\usepackage[dvipsnames]{xcolor}
\usepackage{graphicx,booktabs,tabularx,microtype,refcount}
\usepackage[numbers,sort&compress]{natbib}
\usepackage{colortbl}
\definecolor{trrmethod}{gray}{0.94}
\newcommand{\trrtablelayout}{%
  \renewcommand{\arraystretch}{1.12}%
  \setlength{\heavyrulewidth}{0.65pt}%
  \setlength{\lightrulewidth}{0.35pt}%
  \setlength{\cmidrulewidth}{0.30pt}%
}
\newcommand{\trroutcomes}[3]{%
  \makebox[1.2em][r]{#1}\,{\color{black!45}\textbar}\,%
  \makebox[0.65em][r]{#2}\,{\color{black!45}\textbar}\,%
  \makebox[0.65em][r]{#3}%
}

\usepackage{placeins,tikz,xurl,float}
\usetikzlibrary{positioning,arrows.meta}
\usepackage[font=small,labelfont={bf,sf},labelsep=period,skip=9pt]{caption}
\usepackage{subcaption}
\usepackage{titlesec,fancyhdr,tcolorbox}
\definecolor{trrnavy}{HTML}{17354B}
\definecolor{trraccent}{HTML}{327C88}
\definecolor{trrtint}{HTML}{F3F7F9}
\definecolor{trrmuted}{HTML}{5E707C}
\usepackage[breaklinks,colorlinks,linkcolor=trrnavy,citecolor=trraccent,urlcolor=trraccent]{hyperref}
\usepackage{orcidlink}
\titleformat{\section}{\sffamily\Large\bfseries\color{trrnavy}}{\thesection}{0.65em}{}
\titleformat{\subsection}{\sffamily\large\bfseries\color{trrnavy}}{\thesubsection}{0.65em}{}
\titleformat{\subsubsection}{\sffamily\normalsize\bfseries\color{trrnavy}}{\thesubsubsection}{0.65em}{}
\titlespacing*{\section}{0pt}{19pt plus 3pt}{9pt}
\titlespacing*{\subsection}{0pt}{13pt plus 2pt}{6pt}
\titlespacing*{\subsubsection}{0pt}{10pt}{5pt}
\newcommand{\trrlabheader}{\raisebox{-7pt}{\includegraphics[height=25pt]{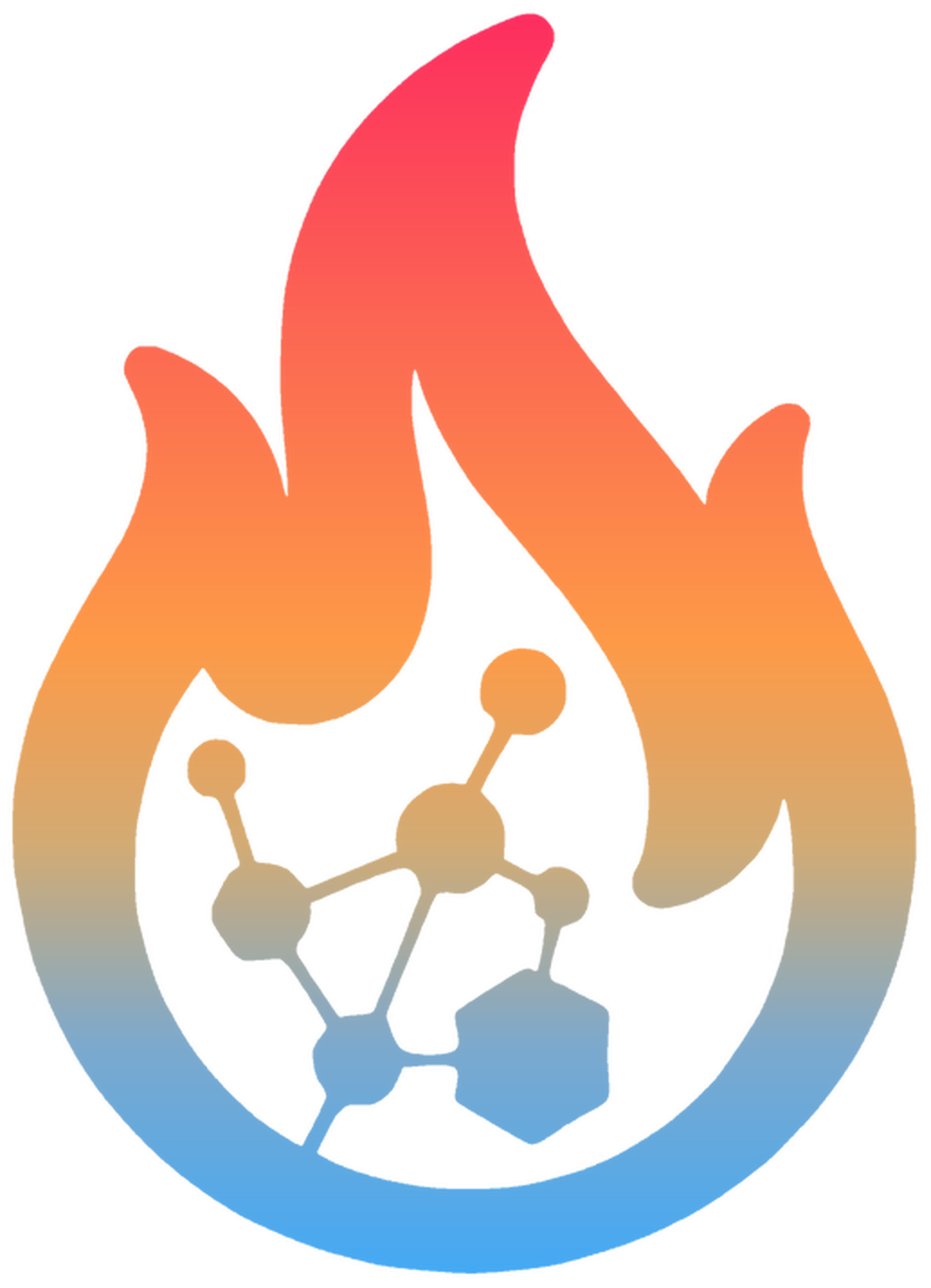}}\hspace{7pt}{\sffamily\bfseries\color{trrnavy}Chizhi Lab}}
\renewcommand{\headrulewidth}{0.5pt}
\renewcommand{\footrulewidth}{0.35pt}
\renewcommand{\headrule}{\hbox to\headwidth{\color{trraccent}\leaders\hrule height \headrulewidth\hfill}}
\renewcommand{\footrule}{\hbox to\headwidth{\color{trrmuted!40}\leaders\hrule height \footrulewidth\hfill}}
\makeatletter
\renewcommand{\maketitle}{%
\vspace*{3pt}\noindent{\sffamily\bfseries\fontsize{35}{38}\selectfont\color{trrnavy}TRR}\par
\vspace{8pt}\noindent{\sffamily\bfseries\fontsize{21}{25}\selectfont\color{trrnavy}A Thermal Interaction Engine for\\Robot Simulation and RGBT Data Generation\par}
\vspace{9pt}\noindent{\sffamily\fontsize{11.5}{15}\selectfont\color{trraccent}Persistent material heat, synchronized RGB--thermal observations and task-grounded robot data.\par}
\vspace{14pt}\noindent\begin{tabular}{@{}l@{}}\@author\end{tabular}\par\vspace{16pt}}
\makeatother
\renewenvironment{abstract}{\begin{tcolorbox}[colback=trrtint,colframe=trraccent,boxrule=0pt,leftrule=2pt,arc=0pt,left=12pt,right=12pt,top=10pt,bottom=10pt]\noindent{\sffamily\bfseries\color{trrnavy}ABSTRACT}\par\vspace{5pt}\small\noindent\ignorespaces}{\end{tcolorbox}\vspace{5pt}}
\newcommand{\trrappendixheading}{\noindent{\sffamily\bfseries\fontsize{25}{29}\selectfont\color{trrnavy}Appendices}\par\vspace{12pt}}
\title{TRR: A Thermal Interaction Engine for Robot Simulation and RGBT Data Generation}
\author{%
Biwen Yang\,\orcidlink{0009-0009-7887-7906}\quad Zhe Cao\quad Jin Zhang\quad Haoyang Sun\\[3pt]
Zonglu Hu\quad Jintao Zhao\quad Zhehao Zhang\quad Kaizheng Chen\\[3pt]
Chengyi Liu\quad Chengyu Wei\quad Siyuan Wu\quad Ruiheng Zhang\,\orcidlink{0000-0002-5460-7196}\textsuperscript{*}\\[6pt]
Beijing Institute of Technology\\[3pt]
{\small $^{*}$Corresponding author}\\[7pt]
{\sffamily\small\href{https://beveryang.github.io/TRR-Thermal-Renderer-for-Robot/}{Project page: beveryang.github.io/TRR-Thermal-Renderer-for-Robot/}}
}
\hypersetup{pdftitle={TRR: A Thermal Interaction Engine for Robot Simulation and RGBT Data Generation},pdfauthor={Biwen Yang, Zhe Cao, Jin Zhang, Haoyang Sun, Zonglu Hu, Jintao Zhao, Zhehao Zhang, Kaizheng Chen, Chengyi Liu, Chengyu Wei, Siyuan Wu, Ruiheng Zhang},pdfsubject={Robot simulation, thermal interaction, RGB-thermal data generation}}

\begin{document}
\maketitle
\begin{abstract}

A robot's actions change the thermal conditions of its next interaction: heat retained after release, carried with objects or stored beneath the surface can change a later contact's outcome despite similar current temperatures. Training trajectories must capture how these effects connect earlier actions to later observations and task outcomes. We present TRR, a \emph{Thermal Robot Renderer} that attaches spatial and depth-resolved thermal state to both contact partners, updates local heat exchange through contact, transport and separation, and generates synchronized RGB--thermal (RGBT) observations and event-local task checks on one execution clock. Controlled interventions show why retained heat, spatial transport and internal storage matter for subsequent contact outcomes. From the same state, TRR produces a paired six-task benchmark of 2,224 static--dynamic pairs, that is, 4,448 trajectories with synchronized dual-view observations, states and actions. Visible thermal history raises safe contact-site selection from 75.0\% to 91.7\% on a dedicated learning corpus, and training on dynamic histories raises simulated transfer completion by 50 percentage points under a new heating procedure with shared feedback execution. We demonstrate RGBT observation adapters on MuJoCo and Isaac Sim, report an engineering inventory of ten prepared robot configurations and six demonstrated camera viewpoints, and describe a developer workflow that carries an existing robot scene to exported RGBT records. The benchmark, the workflow and the platform resources described in Section~\ref{sec:applications} form the first items of a planned public release. Demonstrations and resource updates are available on the project page.

\end{abstract}

\begin{figure*}[!tbp]
\centering
\includegraphics[width=0.94\linewidth]{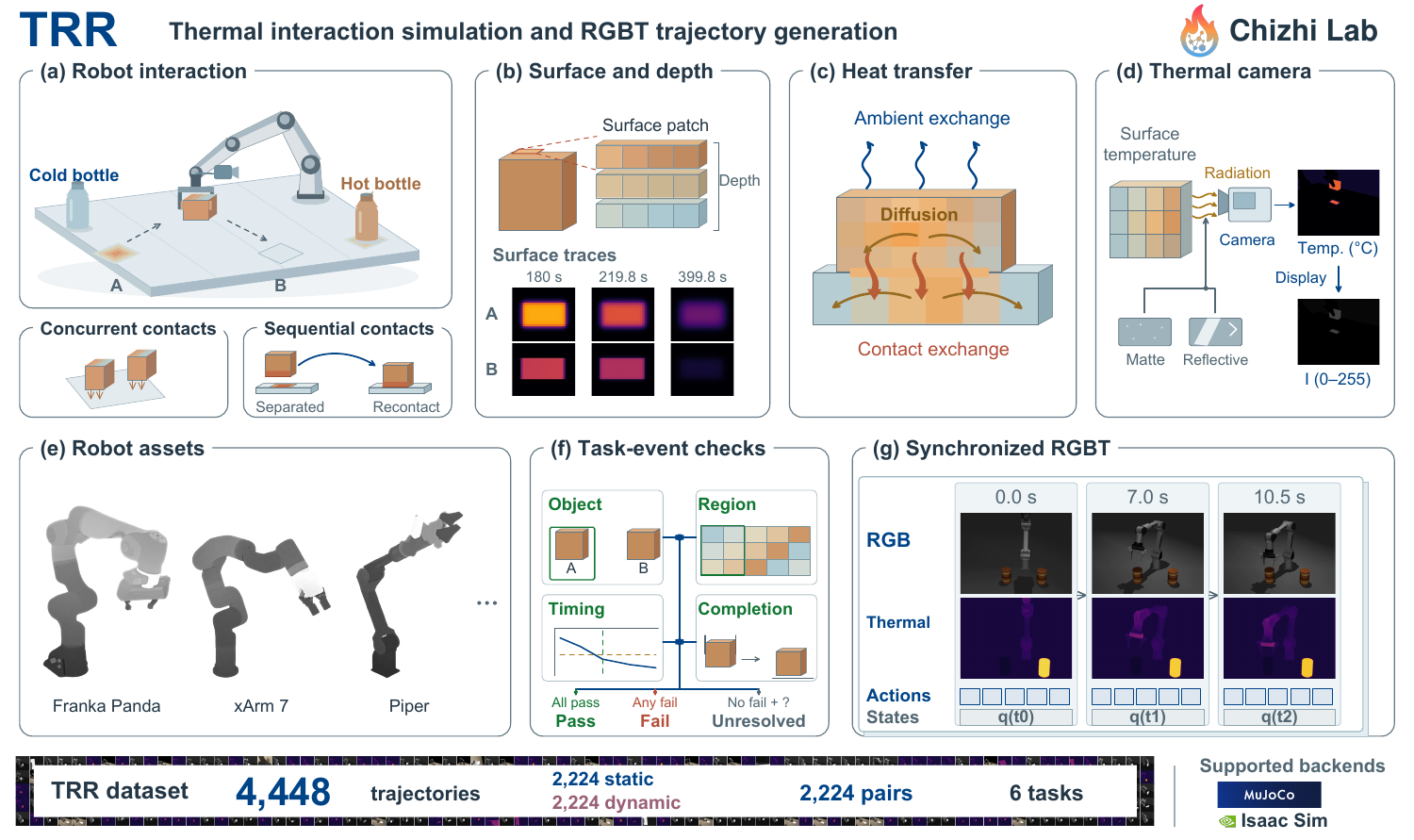}
\caption{\textbf{TRR overview.} Local exchange leaves thermal traces on the contacted surfaces. The evolving fields supply camera observations and task-event checks during the same execution, while each body retains the thermal effects of earlier contacts.}
\label{fig:overview}
\label{fig:architecture}
\end{figure*}

\section{Introduction}
\label{sec:introduction}
A robot lifts a hot container, carries it across a workspace and sets it down. The gripper exchanges heat with the container, the original support retains a thermal trace, and the destination meets a surface altered by the journey. The robot has changed the thermal conditions of its own next action. Thermal manipulation already uses temperature to choose actions \cite{zhou2025motif,son2026thermoact,yu2026safe}, and training data for these decisions has to record how each action changes the conditions and outcomes of later contacts.

States that look equivalent under the current sensor can lead to different outcomes under the same next action. Equal body-mean temperatures hide different local contact histories; nearly identical surface temperatures conceal different internal heat stores. Motion decides which heated region reaches the next interface, and separation leaves heat on the previous partner. A trajectory generator that generalizes across these effects has to carry thermal state on the material itself, through changing contacts and visibility, so that its observations and task labels agree with the state each action meets. For a developer, this is what turns a robot scene into a source of thermally grounded RGB-thermal (RGBT) data: the same frame sequence that shows a robot carry an object also explains why the later contact succeeded or failed.

Existing simulators provide parts of this picture: temperature-dependent household states \cite{li2021igibson,li2023behavior}, heat-related material changes \cite{fu2022rfuniverse}, contact temperature sensing \cite{chung2026tactile}, and external thermal-camera observations \cite{gazebothermal,grimaldi2502stonefish,su2026hearth}. Trajectory generation requires these capabilities to act together across interactions, so that heat left by one contact changes later observations and contact outcomes. Unchanged robot motions then receive different task labels as heat accumulates or dissipates at the contacted region during grasping, transport and subsequent release.

\begin{figure*}[!tbp]
\centering
\includegraphics[trim=0 152.5 0 0,clip,width=0.98\linewidth]{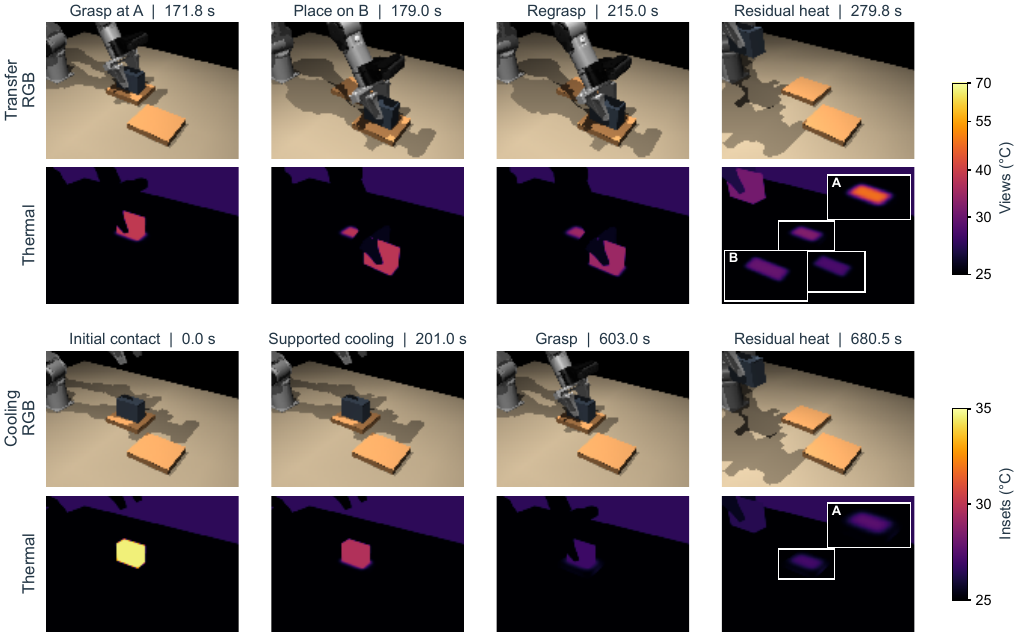}
\caption{\textbf{What the engine records.} RGB and thermal frames from one simulated transfer of a hot aluminium shell (Piper in MuJoCo), taken at the stages of contact, transport, regrasp and release. Thermal frames share a 25--70\,$^{\circ}$C scale, so the warming gripper and support surfaces appear in the same images as the shell. The full sequence, including cooling after separation and residual heat at supports A and B once the shell has left them, appears at a 25--35\,$^{\circ}$C scale in Figure~\ref{fig:process-checks}.}
\label{fig:interaction-sequences}
\end{figure*}

We present TRR, a \emph{Thermal Robot Renderer} built around persistent material-associated thermal state. Object-local thermal cells retain spatial patterns and internal storage as bodies move. Changing contacts couple local regions on both partners through equal-and-opposite heat exchange; separation removes the coupling while preserving both fields. The same evolving state supplies camera observations and event-local task checks on a shared execution clock, and parameterized assets and camera models initialize and calibrate a scene from real measurements. A user gains three things from this design: RGBT trajectories whose labels reflect the thermal conditions each action actually met, paired static--dynamic comparisons that isolate the effect of thermal evolution on a fixed motion, and a labeled view of where in a scene heat is retained, transported or stored.

Section~\ref{sec:related-work} reviews thermal simulation and manipulation data. Section~\ref{sec:method} describes the method, with a plain-terms overview first. Section~\ref{sec:evaluation} reports the interventions, paired benchmark and learning studies. Section~\ref{sec:platform-guide} presents the platform, its developer workflow and a complete MuJoCo use case from scene to exported records, and Section~\ref{sec:applications} the demonstrated uses, the technical and access status of the resources, integration cost and scope, and the outlook. Section~\ref{sec:conclusion} concludes.

The platform derives from the engineering work behind these results. It covers ten robot configurations, two simulation backends, six camera views, measured calibration assets and learning-data interfaces in the recorded October 2026 inventory, and Section~\ref{sec:applications} gives the technical and access status of each resource.

Our contributions are:
\begin{enumerate}
\item \textbf{Persistent spatial and depth-resolved thermal state.} TRR updates material-region states on both sides of a changing contact, retaining local traces and internal heat storage through transport, separation and repeated interaction with newly contacted objects and support surfaces.
\item \textbf{Unified interaction, observation and task checks.} Local heat exchange, material-dependent camera response and event-local checks share the same evolving state and execution clock, linking thermal history to synchronized observations and contact outcomes.
\item \textbf{A paired thermal manipulation benchmark.} TRR produces 2,224 matched static--dynamic condition pairs, that is, 4,448 trajectories across six tasks, with synchronized RGBT observations, states and actions. Dedicated TRR-generated training corpora support safer contact-site choices and a 50-percentage-point gain in simulated transfer completion from dynamic training under a new heating procedure with shared feedback execution and common thermal task criteria.
\item \textbf{A reusable platform workflow.} Reusable assets, backend adapters, a configurable thermal camera and structured records carry the same connection from a robot scene to inspection, learning data and task-level evaluation (Section~\ref{sec:platform-guide}).
\end{enumerate}

\section{Related Work}
\label{sec:related-work}

Simulation platforms provide motion, contact and sensor observations~\cite{todorov2012mujoco,makoviychuk2021isaac,xiang2020sapien,savva2019habitat,zhu2020robosuite,gu2023maniskill2,wang2022tacto,si2024difftactile}. TRR synchronizes contact, field and image queries with thermal state within each executed robot trajectory.

\subsection{Thermal interaction in robot simulators}

Thermal simulation and sensing cover household state and scalar temperatures \cite{li2021igibson,li2023behavior}, decision-relevant evolution \cite{zhou2024hazard}, contact-dependent response~\cite{fu2022rfuniverse,bhattacharjee2015material,bhattacharjee2021material} and fire--radiation coupling \cite{wagner2026fire}. Tactile Genesis diffuses heat in robot-link grids, coupling sensing voxels to a scalar opposing-link temperature \cite{chung2026tactile}; TRR resolves local patches on both partners and ties their surfaces to imaging and task events.

\subsection{Thermal imaging and measured scenes}

Gazebo and Stonefish render thermal observations from assigned temperatures \cite{gazebothermal,grimaldi2502stonefish}; DIRSIG, ThRend and OKTAL-SE add physical scene and sensor models \cite{dirsigthermal,brown2010verification,rs4082477,pedro2021thrend,oktal2022}; digital twins integrate finite-element thermal solvers \cite{kaufmann2023expanding}. PhysIR-Splat models temperature, emissivity, radiation and camera response~\cite{gao2026physir}; neural fields reconstruct thermal views~\cite{hassan2024thermonerf,lin2024thermalnerf,ye2024thermal}, and ETGS models their evolution~\cite{wang2026etgs}; HEARTH maps measured temperatures for fixed-field rollouts \cite{su2026hearth}. Many of these systems supply physically rich thermal or infrared observation for a prescribed temperature field; TRR couples measured initialization, camera modeling and the action-dependent evolution of the observed surface during grasping, transport, separation and recontact.

Visible--infrared datasets and fusion pair observations \cite{hwang2015multispectral,jia2021llvip,shivakumar2020pst900,ha2017mfnet,tang2022piafusion}; manipulation also requires matching observations to contact events. A surface can be occluded during exchange and visible only later; TRR retains the observations and checks the intervening contact.

\subsection{Thermal manipulation and data}

Thermal robotics studies contact and multimodal decisions~\cite{brahmbhatt2019contactdb,brahmbhatt2020contactpose,zhang2023interactive,zhou2025motif,son2026thermoact,yu2026safe,son2026selective}, servoing~\cite{hu2021radiation} and partially observed safety~\cite{kim2025well}. Data-generation systems reuse motion and vary tasks~\cite{mandlekar2023mimicgen,nasiriany2024robocasa,tao2025maniskill3,jiang2025dexmimicgen,xue2025demogen,lin2025constraint,chen2025robotwin}; TRR augments trajectories with thermal validity checks and RGBT labels.

Task- and language-conditioned benchmarks organize goals \cite{9001253,mees2022calvin,liu2023libero}, while recorded datasets add scale \cite{walke2023bridgedata,o2024open,khazatsky2024droid}. Appendix~\ref{app:workflow-comparison} compares broader workflows.

\section{Method}
\label{sec:method}
\label{subsec:system}

\subsection{How TRR works in plain terms}
\label{subsec:method-overview}

TRR keeps a temperature map that belongs to each body rather than to a fixed region of the scene. Every body carries a grid or layered set of thermal cells, each holding a temperature and a heat capacity, and the cells travel with the body as it moves. During an execution, three steps run on every control cycle and a fourth runs at observation times and at contact events:

\begin{enumerate}
\item \textbf{Read the mechanical state.} The backend reports body poses, which surfaces touch, how hard they press, and the elapsed simulation time. Contacts are translated into pairs of material cells, one on each body.
\item \textbf{Exchange heat where bodies touch.} A touching pair of cells swaps energy in proportion to their temperature difference and to the contact area, with equal and opposite amounts added to the two partners, so exchange between the two is conservative. Heat entering from sources and leaving to the environment is accounted for separately, as step 3 describes.
\item \textbf{Spread heat and lose it to the surroundings.} Inside each body, neighbours pass heat along; exposed surfaces exchange heat with the air and radiate to their surroundings. This continues whether or not the bodies still touch, so a warm patch moves, spreads and cools on its own.
\item \textbf{Sample observations and evaluate task events.} At each configured observation time, the robot backend renders the RGB image and TRR renders the thermal image from the current surface temperatures through the material emissivity and the calibrated sensor response. At contact events, a task check reads the cells actually touched at that moment and returns pass, fail or unresolved.
\end{enumerate}

Separating two bodies stops step 2 for that pair and retains both temperature maps for continued evolution, so a later contact anywhere reads the heat left earlier. A scene-specific thermal baseline and camera calibration set the starting map and the mapping to image values from real measurements, and changing these inputs adapts TRR to a new scene without changing the update above.

Steps 1--3 appear formally in Sections~\ref{subsec:interaction-paradigm} and~\ref{subsec:thermal-state}, and step 4 in Sections~\ref{subsec:radiometry} and~\ref{subsec:task-generation}. Figure~\ref{fig:workflow} in Section~\ref{sec:platform-guide} shows the same loop as an engineering workflow, from a robot scene to exported records, and the panel below states that loop in workflow order.

\begin{figure*}[!tbp]
\centering
\begin{tcolorbox}[colback=gray!6,colframe=gray!45,boxrule=0.4pt,arc=1pt,left=8pt,right=8pt,top=6pt,bottom=7pt]
{\sffamily\bfseries\color{trrnavy}Conceptual pseudocode}\hfill{\sffamily\footnotesize\itshape\color{trrmuted}event-level structure of one execution; not executable code}\par
\vspace{4pt}
\begin{minipage}{\linewidth}
\ttfamily\small
\setlength{\parindent}{0pt}\setlength{\parskip}{2pt}
scene $\leftarrow$ robot, task geometry, motion source, camera model\par
thermal $\leftarrow$ materials, contact resistances, initial fields, ambient conditions\par
\textbf{for} each control cycle at time \textrm{$t$}, with backend poses, contacts and elapsed time:\par
\hspace*{2em}\textbf{exchange} heat on contacted cells (step 2);\par
\hspace*{2em}\textbf{advance} conduction, convection and radiation (step 3)\par
\hspace*{2em}\textbf{if} \textrm{$t$} is an observation time: \textbf{record} RGB from the backend,\par
\hspace*{4em}\textrm{$Y_n$} and \textrm{$I_n$} from TRR, and state \textrm{$q_n$}\par
\hspace*{2em}\textbf{for} each contact event at \textrm{$t$}: \textbf{evaluate} the touched cells;\par
\hspace*{4em}\textbf{record} pass, fail or unresolved\par
\textbf{export} observations with their action intervals, events and task label,\par
\hspace*{2em}for inspection, learning data or task evaluation\par
\end{minipage}
\end{tcolorbox}
\caption*{\footnotesize\textbf{Conceptual pseudocode for the workflow of Figure~\ref{fig:workflow}.} The panel names the inputs of each stage and the records a run exports; $t$ is simulation time, $q_n$ the robot state, $Y_n$ the camera temperature matrix and $I_n$ the grayscale image at observation time $t_n$.}
\end{figure*}

\subsection{Interaction-centered design}
\label{subsec:interaction-paradigm}

A task specifies geometry, materials, initial temperatures and mechanical and thermal requirements; controllers or demonstrations supply actions, and the robot backend supplies poses and contacts. TRR maps each interface to material regions on both bodies, updates their heat exchange, and retains both states through motion and separation (Figure~\ref{fig:architecture}). At time $t$, mechanical state $\mathbf x_t$ places the material cells, and active contacts couple their temperatures $\mathbf T_t$. Thermal integration advances over elapsed mechanical time; Section~\ref{subsec:thermal-state} gives the update, Section~\ref{subsec:radiometry} the observation sampling, which runs at each configured observation time, and Section~\ref{subsec:task-generation} the checks, which are evaluated at contact events. This gives a heated region a stable material identity while its world position, partners and observation change.

\begin{figure}[t]
\centering
\includegraphics[width=0.88\linewidth,height=0.31\textheight,keepaspectratio]{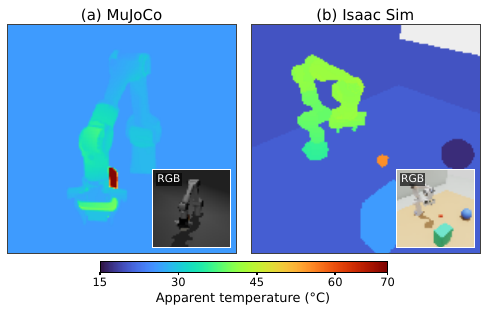}
\caption{RGBT examples from MuJoCo (Piper, left) and Isaac Sim (Franka, right). Temperature views share a scale; insets show same-state RGB. Gray marks unavailable temperature samples.}
\label{fig:backend-rgbt}
\end{figure}

Scene configuration specifies geometry, layered materials, contact resistances, task requirements and camera settings. Real measurements supply a scene-specific thermal baseline and camera calibration. Adapting a scene changes these inputs while retaining the contact mapping and conservation equations below. Figure~\ref{fig:backend-rgbt} illustrates RGB-thermal output from the MuJoCo and Isaac Sim adapters; implementation details for both backends are in Appendix~\ref{app:backend-rgbt}.

\subsection{Local exchange and persistent state}
\label{subsec:thermal-state}

\paragraph{Object-associated material state.}
Cell $i$ stores temperature $T_i$ in kelvin and heat capacity $C_i=\rho_i c_iV_i$, with density $\rho_i$, specific heat $c_i$ and volume $V_i$. Object-local coordinates keep this state attached to the material during motion, and neighbor conductances $G_{ij}$ propagate heat within the body; layered assets resolve near-surface and deeper storage, while finite bulk nodes represent unresolved regions. A world-space contact point $p$ maps to material coordinates $u_o=R_o^\top(p-d_o)$, where $R_o$ and $d_o$ are the object's rotation and translation. Moving the object changes the world positions of its cells rather than resampling their state from a fixed field, so a newly contacted region brings its own temperature and heat capacity into the exchange.

\paragraph{From mechanical contact to local coupling.}
Current poses map backend contacts to surface cells on both bodies. In the transfer and cooling runs, a positive normal force activates eligible contacts. The support footprint comes from projected object geometry, with sample areas accumulated per cell pair and capped by receiving-cell area; the gripper uses a specified pad area. Each interface has a fixed area-normalized resistance, so footprint and temperatures govern exchange during motion (Appendix~\ref{app:contact-configurations}).

\paragraph{Conservative exchange.}
For a local coupling between cells $a$ and $b$, the power entering $a$ is
\begin{equation}
q_{ab}=G_{ab}^{\mathrm{contact}}(T_b-T_a),
\qquad Q_a^{(ab)}=q_{ab},\quad Q_b^{(ab)}=-q_{ab}.
\label{eq:contact-exchange}
\end{equation}
At a layered interface,
\begin{equation}
G_{ab}^{\mathrm{contact}}=
\frac{A_c}{R''_{ab}+d_a/(2k_a)+d_b/(2k_b)},
\label{eq:layered-contact-conductance}
\end{equation}
where $R''_{ab}$ is the area-normalized contact resistance, $A_c$ the represented contact area, and the outer-cell depths $d_a,d_b$ and conductivities $k_a,k_b$ give the two half-cell resistances. Active couplings sum at each cell, and the opposing contributions update both partners while conserving energy.

\paragraph{Propagation.}
The cell state evolves according to
\begin{equation}
\begin{split}
C_i\frac{\mathrm dT_i}{\mathrm dt}
={}&\sum_{j\in\mathcal N_i}G_{ij}(T_j-T_i)
+Q_i^{\mathrm{contact}}+Q_i^{\mathrm{source}}\\
&+h_iA_i^{\mathrm{exp}}(T_{\mathrm{air}}-T_i)+Q_i^{\mathrm{rad}}.
\end{split}
\label{eq:thermal-balance}
\end{equation}
Here $Q_i^{\mathrm{source}}$ is applied heat input, $h_iA_i^{\mathrm{exp}}(T_{\mathrm{air}}-T_i)$ is convection and $Q_i^{\mathrm{rad}}$ is environmental radiative exchange; contact coverage reduces exposed area, and conduction spreads disturbances laterally and through depth (Appendix~\ref{app:thermal-discretization}).

\paragraph{Multiple contacts and initialization.}
The same balance sums support and pad contacts during grasp, lift, placement and release, so changing motion phases changes active couplings without resetting temperatures. A repeated contact at the same pose reads the state left earlier. Separation removes the couplings but preserves both fields, so a later contact elsewhere reads the accumulated state (Section~\ref{subsec:task-generation}). Initialization sets $\mathbf T_0$ from a prescribed or measured baseline; in the reference-camera configuration, registered apparent temperatures supply a surface-temperature estimate and depth assignment sets the stored energy (Appendix~\ref{app:thermal-calibration}).

\subsection{Thermal observations from the evolving state}
\label{subsec:radiometry}
\label{subsec:sensor}
\label{subsec:measured-assets}

\paragraph{Surface readout.}
Camera observations are formed from the current surface temperature. For layered assets, TRR recovers $T^{\mathrm{surf}}$ from the outer-cell state and local boundary heat balance (Eq.~\eqref{eq:surface-boundary-balance}); surface-network cells use their state directly. Object-local interpolation reconstructs the visible field without changing stored energy, and this field supplies both imaging and surface-temperature task queries.

\paragraph{Material and camera response.}
Projection selects a visible surface point $u$. Direct-temperature mode encodes $T^{\mathrm{surf}}(u,t)$ using Eq.~\eqref{eq:direct-temperature-signal}. Material-dependent mode combines emission and reflected surroundings through a broadband graybody approximation:
\begin{equation}
\widetilde L_{\mathrm{surf}}=
\varepsilon(\mathbf v,m)\frac{\sigma[T^{\mathrm{surf}}(u,t)]^4}{\pi}
+[1-\varepsilon(\mathbf v,m)]V\widetilde L_{\mathrm{refl}}(\mathbf v,m,\mathcal E).
\label{eq:equivalent-radiance}
\end{equation}
Here $\varepsilon(\mathbf v,m)$ is emissivity in viewing direction $\mathbf v$, $\sigma$ is the Stefan--Boltzmann constant, and $V\widetilde L_{\mathrm{refl}}$ is visible reflected radiation from surroundings $\mathcal E$. Spatial sampling and sensor response then produce the equivalent radiance $S_{c,t}$ in $\mathrm{W\,m^{-2}\,sr^{-1}}$ for each camera and observation time:
\begin{equation}
\widehat Y^{\mathrm{opt}}_{K,c,t}
=\left(\frac{\pi S_{c,t}}{\sigma}\right)^{1/4},
\qquad \widehat I_{c,t}=\mathcal D_{\phi_c}(\widehat Y^{\mathrm{opt}}_{c,t}),
\label{eq:radiance-readout}
\end{equation}
where $\mathcal D_{\phi_c}$ maps optical temperature to grayscale and the subscript $K$ denotes kelvin. In material-dependent mode, $\widehat Y^{\mathrm{opt}}$ is a blackbody-equivalent temperature and can differ from $T^{\mathrm{surf}}$; Appendix~\ref{app:camera-response} specifies filtering, temporal response, noise and display mapping.

\paragraph{Scene-specific measurement calibration.}
TRR adapts its thermal asset and camera response to real observations. Registered temperature fields set the measured baseline; paired temperature matrices $Y$ and grayscale images $I$ fit the display mapping (Appendices~\ref{app:field-fitting} and~\ref{app:camera-response}). The resulting asset and camera configuration are reused across robot poses. Heat exchange uses material temperatures, while emissivity, reflection and display settings determine camera appearance. This separation allows observation calibration for a capture setup while retaining the same thermal evolution and task checks at the executed contact regions.

\raggedbottom
\setcounter{topnumber}{2}
\setcounter{totalnumber}{4}
\renewcommand{\topfraction}{0.9}
\renewcommand{\textfraction}{0.08}

\subsection{Task-conditioned RGBT generation}
\label{subsec:task-generation}

\paragraph{Matching actions to conditions.}
Each task defines targets and thermal requirements at decision events. A source demonstration or task-specific controller supplies the target and grasp; Cartesian perturbations with inverse kinematics vary the path, and selection variants obtain condition-matched targets from the motion source. Every candidate and replacement is executed and checked against the evolving field. A controller can also use thermal state to change timing: in cooling-and-transfer, a rule-based expert waits and starts grasping when object temperature reaches a threshold; the expert reads physical state while the camera separately generates observations, so the evolving physical state determines timing and the camera records the corresponding appearance during the same executed motion.

\paragraph{Event-local checks.}
Object checks test the selected identity; region checks query the actual contact location at contact time; timing checks specify a region, statistic, threshold and window and require the action within it; mechanical checks assess grasp, transport, release and placement. Each returns pass, fail or unresolved, so an execution is negative if any required check fails, positive if all pass and unresolved otherwise. With $\chi$ indicating a definite pass and $\mathcal K(\tau)$ indexing thermal checks,
\begin{equation}
\operatorname{accept}(\tau)=
\chi_{\mathrm{mech}}(\tau)\land\chi_{\mathrm{obj}}(\tau)\land
\bigwedge_{k\in\mathcal K(\tau)}\chi_k(\tau).
\label{eq:trajectory-acceptance}
\end{equation}
The archive retains all three outcomes with their synchronized observations and actions.

\paragraph{Synchronized records.}
At time $t_n$ a record stores RGB, camera temperature $Y_n$, grayscale $I_n$ and robot state $q_n$, paired with timestamped controls over $[t_n,t_{n+1})$; the terminal observation closes the last transition, preserving high-rate actions between camera frames. A default single-action entry selects the first control target after the observation, and the full interval remains available; static and dynamic records share this convention. Thermal-condition variants form task instances, and additional views or camera-model variants retain episode identity, distinguishing additional observations from independently executed interaction trajectories.

\section{Experiments}
\label{sec:evaluation}
Experiments trace the connection from thermal state to observations, task outcomes and learned decisions. Following analytical contact and diffusion checks (Appendix~\ref{app:task-variations}), interventions ask which state distinctions change the next contact. Measured observations and camera-driven timing test the connection to sensing. Paired trajectories then isolate changes in task validity, and separate learning corpora connect thermal training data to safe contact choices and successful robot transfer under a new heating procedure.

\subsection{Local thermal state across interactions}
\label{subsec:eval-process}\label{subsec:eval-tasks}\label{subsec:mechanism-interventions}\label{subsec:contact-decision}
We begin with a transfer scene containing a hot aluminum shell, two layered supports and finite gripper pads, totaling 10,503 thermal nodes. Ambient and initial shell temperatures are 25 and $70\,^{\circ}$C; moving phases use 500~Hz mechanics and holds use settled contacts. The expert starts its approach at 168.8~s, when the shell reaches $39.993\,^{\circ}$C. Subsequent lifting exposes traces on both supports. Appendix~\ref{app:execution-records-codex041} gives the full execution; the following interventions test which parts of its thermal history change the next contact.

\begin{figure}[!t]
\centering
\includegraphics[width=0.88\linewidth,height=0.31\textheight,keepaspectratio]{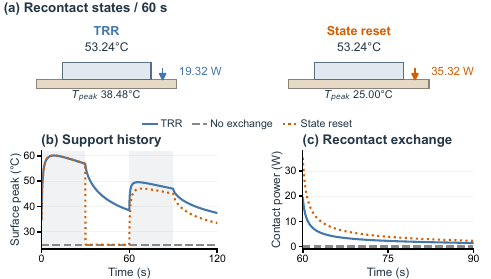}
\caption{Retained support heat changes recontact. The shell has the same precontact temperature in the retained and reset runs, while the receiving surface and contact power differ. Shading marks contact; no exchange keeps the support at ambient.}
\label{fig:mechanism-interventions}
\end{figure}

\paragraph{Exchange and retained support heat.}
A shell contacts support A during 0--30~s, separates during 30--60~s, and returns during 60--90~s. We compare unchanged coupling, zero contact conductance, and resetting only the support to ambient at separation. Without exchange the support stays at $25\,^{\circ}$C. With retained state, its surface peak is $38.48\,^{\circ}$C before recontact. The retained and reset runs have the same shell temperature, $53.24\,^{\circ}$C, at that instant, but initial recontact power is 19.32 versus 35.32~W (Figure~\ref{fig:mechanism-interventions}). By 90~s the shell reaches $47.61$ versus $46.26\,^{\circ}$C. Retaining the support state therefore matters even when the returning object has the same temperature. Appendix~\ref{app:mechanism-protocol} gives the intervention and energy accounting for each contact partner.

\begin{figure}[!t]
\centering
\includegraphics[width=0.76\linewidth]{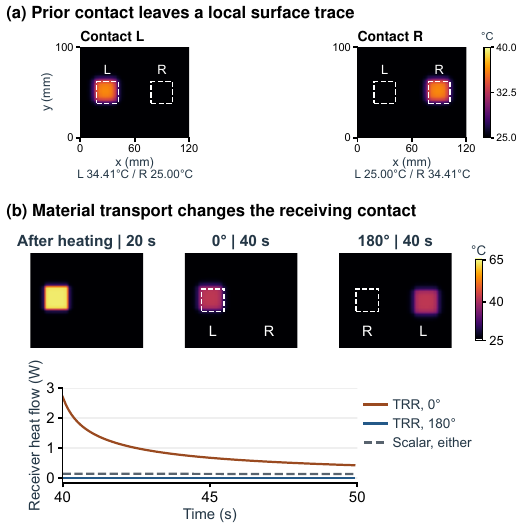}
\caption{Spatial history at a later contact. (a) Prior contact at L or R leaves different local fields despite equal plate-wide means. (b) Heating, translating and rotating a plate changes which material patch reaches the fixed receiver; dashed boxes mark its footprint. The two interventions have their own contact schedules and temperature scales for their respective contact configurations.}
\label{fig:transport-history}
\end{figure}

\paragraph{Spatial state before and after transport.}
We first shift a $24\times24$~mm footprint between L and R on a stationary plate, keeping material, initial temperature, area and 20~s contact duration fixed. After 40~s of separation, patch means are $34.41/25.00\,^{\circ}$C for left contact and $25.00/34.41\,^{\circ}$C for right contact (Figure~\ref{fig:transport-history}a). Both plate-wide means are $25.55\,^{\circ}$C, so the mean cannot identify the cooler patch. A separate ten-history phone probe reproduces this reversal with mean-peak differences at most $0.083\,^{\circ}$C (Table~\ref{tab:same-mean-history-sup}).

The same distinction matters when the material moves. We heat a plate's left patch for 20~s, translate it by 80~mm at either $0^\circ$ or $180^\circ$ yaw, and contact the same receiver. Just before contact, the warm and opposite patches are $39.56$ and $25.00\,^{\circ}$C. Rotation changes initial receiver heat flow from 2.72~W to below 0.001~W; a scalar plate with matched capacity, conductance and exposed area predicts 0.139~W for both orientations (Figure~\ref{fig:transport-history}b). The receiving interface must query the arriving material region, rather than a world-fixed field or body mean that omits the transported heat pattern.

\begin{figure}[!t]
\centering
\includegraphics[width=0.88\linewidth,height=0.31\textheight,keepaspectratio]{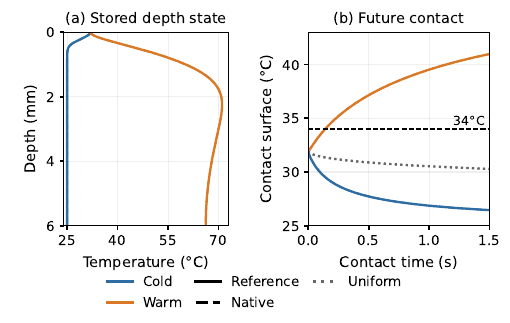}
\caption{Similar current surface, different internal storage. (a) Cold and warm decision-depth profiles. (b) Instantaneous object-side contact temperature under the same finite pad. Reference and native curves nearly overlap. Both surface-matched uniform predictions are shown in gray and nearly coincide; the dashed line marks the $34\,^{\circ}$C contact threshold for both histories.}
\label{fig:hidden-depth-034}
\end{figure}

\paragraph{Internal storage behind a similar surface.}
Two otherwise identical 1~cm$^2$, 6~mm slabs start at $25\,^{\circ}$C. One receives a 0.05~s surface pulse and the other 180~s heating, followed by controlled cooling and 0.1~s natural release. Their reconstructed surfaces differ by only $0.044\,^{\circ}$C ($31.923/31.967\,^{\circ}$C), while stored energies above $25\,^{\circ}$C are $0.261/35.425$~J. Identical finite-pad recontact produces reference peaks of $31.905/40.976\,^{\circ}$C, crossing the common $34\,^{\circ}$C rule only for the warm core (Figure~\ref{fig:hidden-depth-034}). TRR's full layer state reproduces these peaks; a uniform-depth state fitted to each current surface predicts $31.905/31.936\,^{\circ}$C and misses the warm-core violation. The control preserves material, heat capacity and the current surface temperature while replacing the internal depth profile.

A predeclared 16-condition sensitivity grid retains 12 feasible conditions and four infeasible ones. Across the feasible condition--history pairs, full-state and uniform predictions make 0/24 and 12/24 safety errors, respectively. The independent depth reference supplies the future-contact temperatures for these numerical comparisons. Appendix~\ref{app:hidden-depth-protocol} reports the controlled cooling boundary, refinement checks and all conditions; Table~\ref{tab:hidden-depth-034} gives the baseline peaks.

\subsection{Thermal observations and camera-driven execution}
\label{subsec:obs}\label{subsec:eval-observations}
\begin{figure}[!t]
\centering
\includegraphics[width=0.88\linewidth,height=0.31\textheight,keepaspectratio]{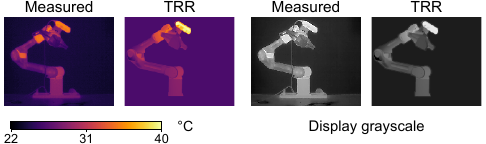}
\caption{Measured and rendered steady-state observations for Piper pose 7. Temperature images share a scale; grayscale uses the calibrated display range for the measured capture setup.}
\label{fig:measured-observations}
\end{figure}

\paragraph{Measured steady-state observations.}
We evaluate scene-specific calibration on ten Piper configurations from one M600 capture session. The measured apparent-temperature field and calibrated display response are held fixed across all ten poses, which are excluded from fitting. Errors are evaluated on native arrays over the supported foreground, giving mean per-pose apparent-temperature MAE $1.442\,^{\circ}$C and RMSE $2.145\,^{\circ}$C. Figure~\ref{fig:measured-observations} shows one pose; Appendix~\ref{app:m600-pose-reuse} gives all ten and the evaluation mask. The temperature agreement demonstrates reuse of the calibrated thermal asset across held-out robot configurations.

\begin{samepage}
\paragraph{Camera-driven transfer in simulation.}
A rule-based controller reads the 2~Hz thermal camera, with $0.15\,^{\circ}$C pixel noise, and starts the same transfer motion at a $40\,^{\circ}$C trigger. At initial shell temperatures of 60, 70 and $80\,^{\circ}$C, raw apparent temperature triggers 17.3--19.2~s early, when the physical temperature is about $41.5\,^{\circ}$C. Correcting the graybody relation using configured emissivity and ambient reflection gives delays of 0.2--0.3~s and physical start temperatures of $39.98\,^{\circ}$C. All nine simulated transfers complete mechanically (Appendix~\ref{app:camera-trigger}). The corrected camera readout initiates motion at the intended physical temperature, within camera-sampling precision at the configured 2~Hz observation rate.
\par\end{samepage}

\subsection{Paired trajectories and thermal task validity}
\label{subsec:eval-data}\label{subsec:eval-scaling}
\begin{figure}[!t]
\centering
\includegraphics[width=0.88\linewidth,height=0.31\textheight,keepaspectratio]{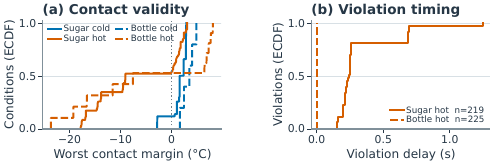}
\caption{Validity and timing of paired thermal tasks. (a) Worst supported margin per condition. (b) First violation delay among detected violations. Dashed curves denote bottle tasks.}
\label{fig:contact-validity}
\end{figure}

\paragraph{Paired archive and evaluation unit.}
The six tasks select a cooler or hotter mug, or interact with cold or hot regions on a bottle or sugar box. A source condition fixes scene, initial field, task, cameras and executable action trace. Its static and dynamic branches share poses, contact schedule and task predicate; only field evolution differs. The archive contains 2,224 pairs, 4,448 trajectories and 447,318 synchronized dual-view instants. Each source condition is one evaluation unit; labels query the actual contact location at native event time. Object identity, mechanical execution and thermal checks retain pass, fail and unresolved outcomes. Timestamped states and native action intervals preserve the connection between observations and execution (Appendix~\ref{app:experimental-details}).
\begin{table}[t]
\trrtablelayout
\centering\small
\caption{Paired benchmark composition. All static trajectories pass; the columns show dynamic outcomes. The archive contains 4,448 trajectories and 447,318 synchronized dual-view instants.}
\label{tab:benchmark-composition}
\setlength{\tabcolsep}{3pt}
\begin{tabularx}{\linewidth}{@{}Xrrrr@{}}
\toprule
{\bfseries Task} & {\bfseries Pairs} & {\bfseries Pass} & {\bfseries Fail} & {\bfseries Unresolved} \\
\midrule
Cooler mug & 555 & 555 & 0 & 0 \\
Hotter mug & 300 & 300 & 0 & 0 \\
Bottle, cold & 225 & 225 & 0 & 0 \\
Bottle, hot & 425 & 200 & 225 & 0 \\
Sugar box, cold & 300 & 264 & 0 & 36 \\
Sugar box, hot & 419 & 191 & 219 & 9 \\
\midrule
\textbf{Total} & \textbf{2,224} & \textbf{1,735} & \textbf{444} & \textbf{45} \\
\bottomrule
\end{tabularx}
\end{table}
\paragraph{Thermal changes in otherwise matched trajectories.}
Evolving fields turn 444 fixed-field passes into violations and 45 into unresolved outcomes (Table~\ref{tab:benchmark-composition}); violations occur in the two hot-region tasks. Figure~\ref{fig:contact-validity} uses the worst supported margin to the task threshold: intervals wholly on one side pass or fail, while unsupported or threshold-crossing intervals are unresolved. All 225 bottle violations occur at first contact. The 219 sugar-box violations appear 0.15--1.25~s later (median 0.24~s); 53 initially pass. Checking throughout contact captures both immediate and delayed violations in the archive using the same event-local task criteria.

\paragraph{Precontact images forecast delayed violations.}
To test whether these changes are visible before contact, a separate controlled probe reads contact-centered crops at 0.60 and 0.10~s before contact and predicts a drop below $34\,^{\circ}$C within 1.5~s among initially valid contacts. On 137 held-out trials from 24 groups, average precision is 0.261 with RGB, 0.337 with frozen-field RGBT and 0.939 with evolving-field RGBT. Early and late single thermal frames perform similarly to the pair (Appendix~\ref{app:forecast}). At a fixed contact point, one thermal image can already expose the state left by prior interaction. The next study asks a different question: selecting between candidate contacts when the decision view can hide the state left by the preceding contact history.

\subsection{Thermal data for decisions and execution}
\label{subsec:dynamic-decision}
\paragraph{Controlled learning protocol.}
A separate corpus of flat material samples (coupons) contains 240 source groups across four contact-history families. Each decision supplies four RGBT frames and two candidate sites; the occluded family hides the final view. An independent finite-difference reference certifies the contact-surface outcome. The primary test has 128 decisions in 64 held-out source groups, with all splits grouped before training. Every learned arm uses the same 256 source episodes and 40 training epochs, evaluated at epoch 40 for seeds 0--4. The full image-to-site policy uses a ResNet-18 encoder and GRU. The action head supports site and wait; the illustrated contact choices use wait~0. Appendix~\ref{app:dynamic-decision-protocol} gives the full data roles and training settings.

\begin{figure}[H]
\centering
\includegraphics[width=0.94\textwidth,height=0.34\textheight,keepaspectratio]{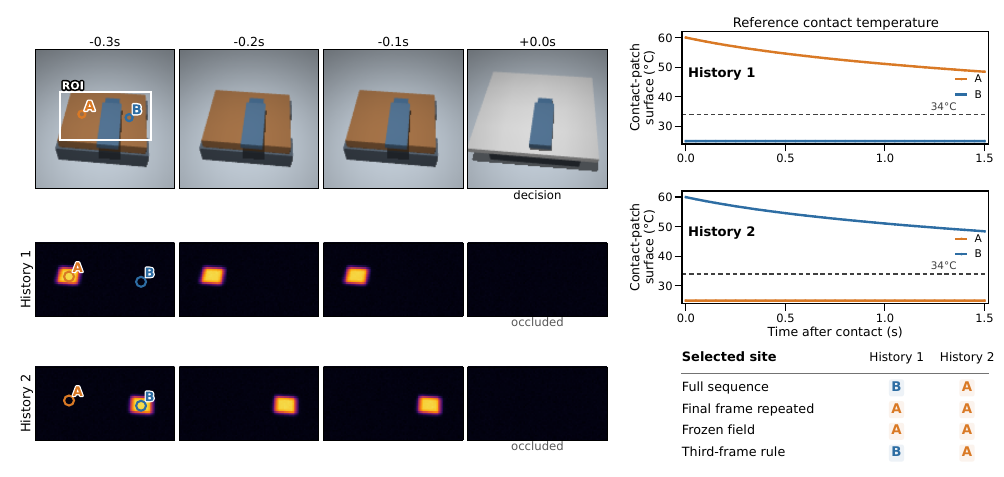}
\caption{Thermal history resolves an occluded contact choice. Left: shared RGB and thermal crops for two histories (24--58\,$^{\circ}$C), timed relative to the decision; the shutter hides the final view. Right: independent contact-surface temperatures, timed from contact, and seed-0 choices. Contact begins 0.18~s after the decision; the dashed line is 34\,$^{\circ}$C. A/B are sites 0/1 at wait~0.}
\label{fig:round10-case}
\end{figure}

\begin{samepage}
The final-frame-repeat control uses the same encoder, GRU and action head, repeating the decision frame four times. Frozen input and spatial means replace observations; RGB-only changes channels, and the global-mean GRU removes the spatial encoder. A fixed third-frame crop rule provides a strong nonlearned comparison. Figure~\ref{fig:round10-case} shows how visible history changes the safe site, and Table~\ref{tab:dynamic-decision} reports the complete primary comparison.
\par\end{samepage}

\begin{table}[H]
\trrtablelayout
\centering\small
\caption{Image-to-site results on 128 held-out decisions (64 groups), averaged over seeds 0--4. Safe is the percentage of certified-safe choices; site match is exact agreement with the expert site. Bold and underline mark the highest and second-highest rates.}
\label{tab:dynamic-decision}
\setlength{\tabcolsep}{4pt}
\begin{tabularx}{\columnwidth}{Xrr}
\toprule
{\bfseries Input / model} & {\bfseries Safe (\%)} & {\bfseries Site match (\%)} \\
\midrule
\rowcolor{trrmethod}Full RGBT sequence & \textbf{91.7} & \textbf{80.8} \\
Final-frame repeat & 75.0 & 62.7 \\
Frozen input & 75.0 & 62.7 \\
Per-frame spatial mean & 72.7 & 60.2 \\
RGB only & 72.5 & 60.0 \\
Global-mean GRU & 62.5 & 50.0 \\
\midrule
Third-frame rule (untrained) & \underline{85.2} & \underline{74.2} \\
\bottomrule
\end{tabularx}
\end{table}
\paragraph{Safe choices and the information in visible history.}
The full sequence scores 91.7\% safe and 80.8\% site match. Final-frame repeat scores 75.0\% and 62.7\%, respectively. The paired source-group safe-choice difference is 16.72 points (95\% group bootstrap interval [11.56,22.03]). Frozen input has the same observed rate; spatial means, RGB-only and global means are lower. The third-frame rule scores 85.2\% safe and 74.2\% site match. The learned policy is 6.56 points higher (95\% group interval [3.43,10.31]). Both use the visible thermal contrast to recover the illustrated site reversal between the two preceding contact histories.

Channel controls are detailed in Appendix~\ref{app:channel-study}. This study tests the information in visible history; the next tests the effect of thermal training data on completed manipulation using learned regions and a shared feedback executor.

\paragraph{From region choice to executed transfer.}
A MuJoCo manipulation study tests whether thermal training data improves transfer completion (Appendix~\ref{app:native-modular055}). An infrared classifier selects one of two bottle regions; the same feedback executor then grasps, transfers and releases the object in MuJoCo with finite bilateral heat exchange. Training on dynamic histories (D) is compared with unheated precontact states (S), with latest-visible-frame repeat using D's labels as an observation control (C). All learned arms share the architecture and three initializations, evaluating only the final weights after 1,000 updates. D and S receive the same dynamic observations at test. Both tests use object placements excluded from training. The first retains the training heating procedure; the second uses heating, quenching and opposite-region reheating on the same bottle. Each has 16 test conditions for paired evaluation of all learned arms under shared task criteria.

\begingroup
\let\trroriginaltable\table
\renewcommand{\table}[1][]{\trroriginaltable[H]}
\begin{table}[t]
\trrtablelayout
\centering\small
\caption{Transfer outcomes per 16 conditions. Cells list completed, failed and unresolved runs, in that order. Seeds are learner initializations; all test placements are held out.}
\label{tab:native-modular055}
\setlength{\tabcolsep}{3pt}
\begin{tabularx}{\columnwidth}{Xccc}
\toprule
{\bfseries Training input} & {\bfseries Seed} & {\bfseries \shortstack{Same heating\\procedure}} & {\bfseries \shortstack{New heating\\procedure}} \\
\midrule
\rowcolor{trrmethod}Dynamic (D) & 0--2 & \trroutcomes{12}{2}{2} & \trroutcomes{13}{1}{2} \\
Static (S) & 0--2 & \trroutcomes{6}{9}{1} & \trroutcomes{5}{9}{2} \\
Visible-frame repeat (C) & 0, 1 & \trroutcomes{12}{2}{2} & \trroutcomes{13}{1}{2} \\
Visible-frame repeat (C) & 2 & \trroutcomes{12}{2}{2} & \trroutcomes{11}{3}{2} \\
\bottomrule
\end{tabularx}
\end{table}

\endgroup
With the training heating procedure, D certifies 12 completions versus six for S per 16 conditions, with a paired difference of $[37.5,43.75]$ percentage points (Table~\ref{tab:native-modular055}).

\begin{figure}[!t]
\centering
\includegraphics[width=0.88\linewidth,height=0.31\textheight,keepaspectratio]{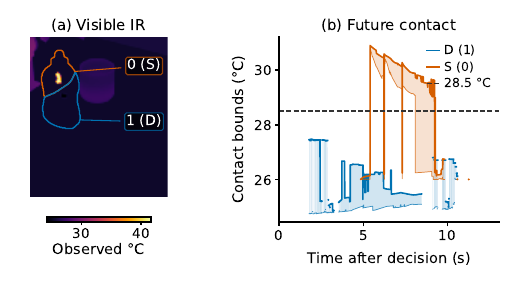}
\caption{Simulated transfer under a new heating procedure. (a) Latest visible IR and D/S region choices. (b) Lower/upper bounds on the maximum qualifying-contact point temperature; gaps indicate no qualifying contact. D passes; S violates the 28.5\,$^{\circ}$C limit. Callouts mark the candidate regions.}
\label{fig:native-modular055}
\end{figure}

Under the new heating procedure, D has 13 certified completions, one failure and two unresolved outcomes per 16 conditions for every training seed; S has five, nine and two. Eight confirmed unilateral wins and no losses give a paired difference of 50 percentage points: the two unresolved pairs use the same physical action and cancel. Absolute task-success fractions remain bounded by $[81.25,93.75]\%$ and $[31.25,43.75]\%$, respectively. C has 13, 13 and 11 certified completions across the three initializations. Figure~\ref{fig:native-modular055} shows an example selected from model choices before execution.

Under the new heating procedure, seven gains avoid thermal violations and the eighth reflects primitive reachability.

\paragraph{From thermal interaction to training signals.}
The interventions expose a common failure of compressed thermal state: equal body means or similar current surfaces can conceal different future exchanges. Retained heat changes the receiving surface's response, transport brings a different material region to the interface, and internal storage sustains heat after surface cooling. Preserving these distinctions gives trajectory generation a physical basis for its labels. In the paired archive, 444 identical motions change from pass to violation when local heat evolves during execution.

\begin{samepage}
TRR records both what happened during contact and what became visible afterward. Timestamped action intervals preserve exchanges between camera frames, and event-local checks associate those exchanges with their task outcomes. For a fixed executed trajectory, varying the camera response changes observations while preserving thermal evolution and contact outcomes. The learning studies connect these records to action: visible history supports contact-site selection, and dynamic training improves transfer with shared feedback execution. The latter comparison holds test observations and the executor fixed, linking different region choices to their thermal consequences throughout grasping and transfer. This gives generated data a causal structure: earlier actions alter later contact conditions, and the recorded observations and task outcomes expose those changes to downstream learning.
\par\end{samepage}

\section{Platform and Developer Workflows}
\label{sec:platform-guide}

TRR brings thermal assets, robot simulation, camera observations and trajectory records into one development workflow. The paired benchmark of Section~\ref{sec:evaluation} is one product of it, and the same route produces the calibration studies, demonstrations and downstream learning data reported here. Figure~\ref{fig:workflow} and Section~\ref{subsec:dev-workflow} state what a developer supplies and receives, and the end-to-end case of Section~\ref{subsec:usecase} follows one execution from scene to exported records. The remaining subsections describe the reusable resources a developer draws on: the inventory (Section~\ref{subsec:assets}), calibration and asset reuse (Section~\ref{subsec:calibration}), backend adapters and camera configuration (Section~\ref{subsec:backend}), and the record formats for demonstrations and learning data (Section~\ref{subsec:records}).

\subsection{Developer workflow}
\label{subsec:dev-workflow}

\begin{figure}[H]
\centering
\resizebox{0.96\linewidth}{!}{%
\begin{tikzpicture}[font=\sffamily\small,
 stage/.style={draw=trraccent,fill=trrtint,rounded corners=2pt,text width=28mm,minimum height=23mm,align=left,inner sep=4pt},
 stream/.style={draw=trrnavy!65,rounded corners=2pt,text width=38mm,minimum height=23mm,align=left,inner sep=4pt},
 flow/.style={-{Latex[length=2mm]},draw=trrnavy,line width=.8pt}]
\node[stage] (scene) at (0,0) {\textbf{1. Scene}\\Robot geometry, task and motion source};
\node[stage] (config) at (40mm,0) {\textbf{2. Configure}\\Materials, initial field and camera};
\node[stage] (run) at (80mm,0) {\textbf{3. Execute}\\Coupled thermal interaction};
\node[stage] (export) at (120mm,0) {\textbf{4. Export}\\Inspect records, build learning data};
\draw[flow] (scene)--(config);\draw[flow] (config)--(run);\draw[flow] (run)--(export);
\node[stream] (obs) at (20mm,-48mm) {\textbf{Observations}\\Backend RGB; TRR temperature and grayscale};
\node[stream] (act) at (70mm,-48mm) {\textbf{Actions and states}\\Timestamped controls and robot state};
\node[stream] (evt) at (120mm,-48mm) {\textbf{Events and labels}\\Contact checks and task outcomes};
\draw[flow] (run.south)--++(0,-8mm)-|(obs.north);
\draw[flow] (run.south)--++(0,-8mm)-|(act.north);
\draw[flow] (run.south)--++(0,-8mm)-|(evt.north);
\draw[trraccent,line width=.7pt] (obs.south)--(20mm,-68mm)--(70mm,-68mm)--(act.south);
\draw[trraccent,line width=.7pt] (act.south)--(70mm,-68mm)--(120mm,-68mm)--(evt.south);
\draw[flow,trraccent] (evt.south)--(120mm,-68mm)--(150mm,-68mm)--(150mm,0)--(export.east);
\end{tikzpicture}}
\caption{\textbf{TRR developer workflow.} A robot scene is configured with thermal assets and a camera model, executed with coupled thermal interaction, and exported. Execute writes three parallel record streams and each stream feeds Export; the streams are synchronized by the same execution and are not derived from one another.}
\label{fig:workflow}
\end{figure}
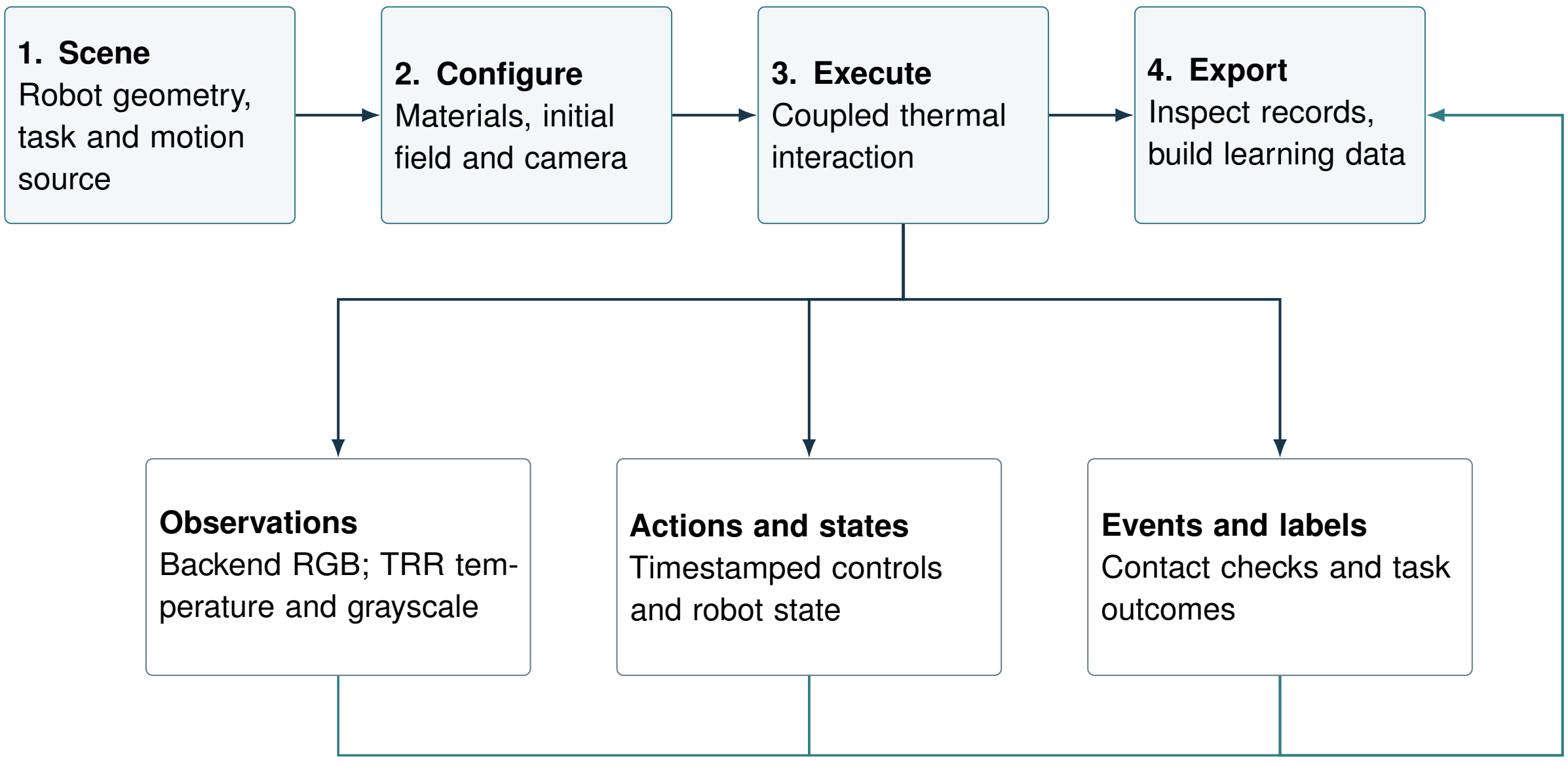

Figure~\ref{fig:workflow} shows the workflow the platform supports, with the inputs and outputs a developer controls and the records a run returns.

\subsection{An end-to-end MuJoCo use case}
\label{subsec:usecase}

The transfer execution of Appendix~\ref{app:execution-records-codex041} runs on this platform, so we follow it end to end. It also appears in Figure~\ref{fig:backend-rgbt} and Figure~\ref{fig:process-checks}.

\paragraph{Input.} A Piper arm in MuJoCo handles a hot aluminium shell at $70\,^{\circ}$C with two layered supports and two finite gripper pads, in a $25\,^{\circ}$C environment and a 10,503-node thermal network. Layered materials set heat capacity and contact resistance, initial temperatures set the shell and support fields, and one external camera sets the equivalent observation model and its display range.

\paragraph{Execution.} The controller waits for the shell to cool to $40\,^{\circ}$C. At the step before 168.8~s the shell reads $39.993\,^{\circ}$C, so the expert starts its approach there, grasps and lifts the shell, which uncovers support A at 174.8~s, transfers it and later exposes the destination trace at 217.3~s. The motion runs at 500~Hz. At each step the backend reports contacts, TRR maps each footprint to cells on both bodies, and conservative exchange, lateral conduction and convective cooling advance the thermal state on the mechanical clock, while the camera samples the configured observation times and the checks read the contacted cells.

\paragraph{Output.} One continuous run preserves the per-channel contact durations and energies for both supports and both pads, the shell, support and gripper histories of Figure~\ref{fig:interaction-curves}, and the exposure of the source and destination traces at 174.8 and 217.3~s. Record volume follows the sampled observation schedule rather than the control rate: the two continuous runs of Appendix~\ref{app:corpus} together save 705 camera samples, and frame--state alignment and finite-array checks cover all of them. Table~\ref{tab:record-fields} lists the record content.

\begin{table}[H]
\centering\small
\caption{Content of a TRR run record. Fields follow the record convention described in Section~\ref{subsec:task-generation}.}
\label{tab:record-fields}
\renewcommand{\arraystretch}{1.16}
\begin{tabularx}{\linewidth}{@{}>{\raggedright\arraybackslash}p{0.20\linewidth}>{\raggedright\arraybackslash}p{0.245\linewidth}>{\raggedright\arraybackslash}X@{}}
\toprule
\textbf{Record} & \textbf{Key fields} & \textbf{Content} \\
\midrule
Observation & observation time, camera id, view & RGB frame, camera temperature matrix $Y_n$, grayscale image $I_n$ \\
\addlinespace[2pt]
Robot state & observation time & robot state $q_n$ \\
\addlinespace[2pt]
Action interval & start and end time & first control target after the observation, with the full native interval retained \\
\addlinespace[2pt]
Event check & event time, check type, region & object, region, timing or mechanical outcome: pass, fail or unresolved \\
\addlinespace[2pt]
Trajectory & trajectory id, task, program & task label from the checks, contact channels and durations \\
\bottomrule
\end{tabularx}
\end{table}

The exported trajectory holds synchronized RGB and thermal evidence of how the shell and both supports changed during one transfer, the actions that produced those changes, and a task label computed from the contact that occurred. It can be compared with its static counterpart, reviewed as a demonstration, or loaded through a learning interface. Section~\ref{sec:applications} describes the demonstrated uses of such records.

\subsection{Reusable assets and scene building blocks}
\label{subsec:assets}

The asset collection separates robot configurations, thermal caches and measured calibration assets. A robot configuration supplies the articulated scene; its thermal cache associates a spatial field with the asset; a measured calibration asset additionally incorporates observations from a physical setup. These resources can be combined with task objects, environmental surfaces and camera placements to construct a demonstration or a data-generation scene.

Table~\ref{tab:platform-assets} summarizes the engineering inventory recorded for the October platform demonstration. It documents recorded resource types alongside their intended roles. The paired benchmark of Section~\ref{sec:evaluation} is a separate data product containing 2,224 source-condition pairs and 4,448 trajectories; its scale is reported independently of the earlier demonstration batches.

\begin{table}[H]
\centering\small
\caption{Engineering resource inventory. Asset and interface counts were recorded on October 12, 2026; thermal-process demonstrations were updated on October 14. Counts describe the corresponding resource type.}
\label{tab:platform-assets}
\renewcommand{\arraystretch}{1.18}
\begin{tabular}{@{}p{0.24\linewidth}p{0.19\linewidth}p{0.50\linewidth}@{}}
\toprule
\textbf{Resource} & \textbf{Inventory} & \textbf{Role and examples} \\
\midrule
Robot configurations & 10 configurations & Articulated scene setups, including Piper, Panda, FR3 and xArm7. \\
Thermal caches & 16 categories & Nine robot caches, four object caches and three environment caches. \\
Measured robot asset & One Piper asset & Piper with D435 camera and mounting bracket; demonstrated reuse across ten poses. \\
Simulation adapters & Two backends & MuJoCo and Isaac Sim, with RGB and thermal observation examples. \\
Task scene families & Three families & Cup, sugar box and bottle, covering six hot/cold task conditions. \\
Camera placements & Six demonstrated views & Three external and three wrist-mounted viewpoints with RGBT output. \\
Camera calibration & One M600 model & Calibrated equivalent observation model; HIK and IRAY devices were also connected for acquisition. \\
Learning-data interfaces & Two formats & LeRobot and robomimic, with small-batch loading checks. \\
Thermal demonstrations & 15 engineering demonstrations & Twelve task replays and three continuous-process demonstrations. \\
\bottomrule
\end{tabular}
\end{table}

\subsection{Scene-specific calibration and asset reuse}
\label{subsec:calibration}

Parameterization lets the same thermal representation serve different scene configurations. Material properties and initial temperature distributions specify the physical state, while camera parameters specify how that state appears in an observation. This separation gives developers distinct controls for changing an interaction and changing its observation. A scene can therefore be adapted through its configuration while retaining the underlying state update and record generation.

The Piper example illustrates a practical calibration route. Measured thermal images, the robot geometry and camera alignment are used to construct a spatial thermal asset and fit the observation response. The resulting field is attached to the robot's material coordinates, allowing articulated pose changes to reuse it. The earlier demonstration includes compact, extended and raised configurations; the measured comparisons and fitting procedure are provided in the experimental sections and appendices. For engineering use, this provides a concrete path from a measured setup to a reusable scene asset.

\subsection{Backend integration and configurable observations}
\label{subsec:backend}

The backend adapter connects scene geometry, body poses, contact information and camera configuration to the thermal representation. The robot simulator supplies mechanical evolution; TRR evolves the corresponding thermal state and renders observations at the configured camera times. MuJoCo and Isaac Sim examples in Figure~\ref{fig:backend-rgbt} show this observation interface on different robot scenes; the two adapters cover RGB and thermal observation, while the dynamic interaction studies of Section~\ref{sec:evaluation} run in MuJoCo.

Camera placement and response are separate configuration choices. External views can show the whole task, while wrist views follow the local interaction. The recorded temperature arrays support numerical processing; rendered thermal images support inspection and camera-response studies. RGB, thermal observations and execution records retain their temporal association, so a developer can inspect the image that accompanies a particular contact or action interval.

\subsection{From execution records to demonstrations and learning data}
\label{subsec:records}

Records reach downstream use through three routes: a contact trace shows when and where exchange occurred, thermal observations show its visible consequences, and task checks connect those consequences to the selected object and interaction interval. For demonstrations, the existing material covers concurrent contacts, transfer between supports, cooling after separation and later recontact. For data generation, the paired static--dynamic protocol holds the mechanical trajectory fixed while changing thermal evolution. Downstream packaging then selects the observations, robot states and actions required by the learning interface; the LeRobot and robomimic loading checks provide concrete examples of this last step.

Together, the workflow diagram and its case, the asset inventory, the configurable thermal and camera models, the backend adapters and the structured records give three entry points into TRR: a robotics developer can build a thermally instrumented scene; a perception researcher can study observations across views and camera settings; and a learning researcher can use generated trajectories or decision windows. All three use the same connection between interaction, evolving thermal state and recorded observations that underlies the research results.

\section{Applications, Availability and Outlook}
\label{sec:applications}

The engine supports research and engineering uses that share one requirement: the thermal condition of a contact must follow from what happened earlier in the same execution. This section states which uses are demonstrated in this paper, which resources currently exist and in what form, and where the work goes next.

\subsection{Demonstrated uses}
\label{subsec:applications-demonstrated}

\begin{table}[H]
\centering\small
\caption{What the research results mean for a user of TRR. Each row names a demonstrated capability and the use it enables.}
\label{tab:applications}
\renewcommand{\arraystretch}{1.16}
\begin{tabularx}{\linewidth}{@{}>{\raggedright\arraybackslash}p{0.315\linewidth}>{\raggedright\arraybackslash}X@{}}
\toprule
\textbf{Demonstrated capability} & \textbf{Use it enables} \\
\midrule
Heat is retained on both contact partners through separation & Study how repeated grasping and recontact change object and support temperatures, and why an equally warm object can meet a differently warm surface. \\
\addlinespace[2pt]
Material regions move with the object & Simulate the local thermal condition of different grasp sites and predict which arriving region meets the next interface. \\
\addlinespace[2pt]
Internal storage behind a similar surface & Treat the visible surface as incomplete evidence: two objects at the same surface temperature can lead to different contact peaks and different safety outcomes. \\
\addlinespace[2pt]
Identical motions with different task labels & Construct and analyze data under dynamic thermal constraints; 444 of 2,224 paired motions move from pass to violation as heat evolves. \\
\addlinespace[2pt]
Thermal history improves contact-site choice & Quantify the value of temporal thermal observations for manipulation decisions, with a matched final-frame control. \\
\addlinespace[2pt]
Dynamic data improves simulated transfer & Provide evidence that dynamic thermal training data changes execution outcomes under a shared feedback executor. \\
\bottomrule
\end{tabularx}
\end{table}

The learning results describe two dedicated corpora. The 75.0\% to 91.7\% contact-site result comes from 240 coupon source groups with a matched final-frame control, and the 50-point transfer result compares dynamic-history training with unheated precontact training in a MuJoCo transfer study, where the latest-visible-frame control with dynamic labels also completes many transfers. The paired benchmark measures a different property: on 2,224 paired motions, 444 pass in a fixed field and violate the task rule once heat evolves. Together they show where dynamic thermal data changes decisions and where a simpler observation suffices.

\subsection{Research and integration use cases}
\label{subsec:applications-cases}

The demonstrated uses support two kinds of engagement.

\begin{itemize}
\item \textbf{Research use.} Reproduce the interventions of Section~\ref{sec:evaluation}, compare thermal training data with static or frozen-field data on a new decision task, or analyze the paired benchmark under dynamic thermal constraints. A researcher can use the paired static--dynamic design as a controlled protocol and substitute a different policy or executor.
\item \textbf{Integration use.} Add thermal interaction to an existing MuJoCo or Isaac Sim robot scene and produce RGBT trajectories for that scene, as in the use case of Section~\ref{subsec:usecase}. A robotics group gains labelled thermal data for its own objects, supports and camera placements without rebuilding the mechanical simulation.
\end{itemize}

A project site will present two entry points that share one engine: thermal simulation integration for teams that already run a robot simulator and need contact exchange, material thermal state and thermal-camera observations in it, and RGBT data generation for learning teams that need synchronized RGBT--action trajectories, event labels and dynamic thermal task data. Both rest on capabilities demonstrated here; neither is offered as a finished product.

\subsection{Availability}
\label{subsec:availability}

Table~\ref{tab:resource-status} separates two questions that the earlier inventory combined. \emph{Technical status} records how far a resource has been verified: demonstrated means used in this paper, reported study marks a learning corpus whose results appear in Section~\ref{subsec:dynamic-decision}, and prepared means an internal working resource with the usage described here. \emph{Access status} records what an external reader can obtain now; the research resources listed below are currently internal, and the release is in preparation as the package organized in Appendix~\ref{app:release}. The \href{https://beveryang.github.io/TRR-Thermal-Renderer-for-Robot/}{TRR project page} provides public demonstrations and resource updates, with presentation assets in the \href{https://github.com/BeverYang/TRR-Thermal-Renderer-for-Robot}{project repository}. Simulator code and research data remain in release preparation.

\begin{table}[H]
\centering\small
\caption{Resource status, with technical status separated from access status. Platform resources follow the October 2026 inventory of Table~\ref{tab:platform-assets}; the benchmark and learning corpora are reported with their own results. Research resources retain their own access status; public preview media are listed separately.}
\label{tab:resource-status}
\renewcommand{\arraystretch}{1.16}
\begin{tabularx}{\linewidth}{@{}>{\raggedright\arraybackslash}p{0.20\linewidth}>{\raggedright\arraybackslash}p{0.125\linewidth}>{\raggedright\arraybackslash}p{0.12\linewidth}>{\raggedright\arraybackslash}X@{}}
\toprule
\textbf{Resource} & \textbf{Technical status} & \textbf{Access status} & \textbf{Content and supporting evidence} \\
\midrule
Project demonstrations & Published & Public & Preview GIF and overview images on the project page and repository. \\
\addlinespace[2pt]
Paired RGBT benchmark & Demonstrated & Internal & 2,224 source-condition pairs, 4,448 trajectories and 447,318 synchronized dual-view instants across six tasks (Table~\ref{tab:benchmark-composition}). \\
\addlinespace[2pt]
Contact-site learning corpus & Reported study & Internal & 240 source groups in four contact-history families; primary comparison on 128 held-out decisions in 64 groups. \\
\addlinespace[2pt]
Regional-decision transfer corpus & Reported study & Internal & Regional choices with shared feedback execution, 16 test conditions per heating track and three initializations. \\
\addlinespace[2pt]
Simulation adapters & Demonstrated & Internal & MuJoCo and Isaac Sim RGBT output (Figure~\ref{fig:backend-rgbt}). \\
\addlinespace[2pt]
Robot configurations & Prepared & Internal & Ten configurations, including Piper, Panda, FR3 and xArm7. \\
\addlinespace[2pt]
Thermal caches & Prepared & Internal & 16 categories: nine robot, four object and three environment caches. \\
\addlinespace[2pt]
Measured Piper asset & Demonstrated & Internal & Calibrated steady-state field reused across ten held-out poses; mean apparent-temperature MAE 1.442\,$^{\circ}$C. \\
\addlinespace[2pt]
Camera model and placements & Demonstrated & Internal & One calibrated M600 equivalent model; six demonstrated views, three external and three wrist-mounted. \\
\addlinespace[2pt]
Thermal demonstrations & Prepared & Internal & 15 engineering demonstrations: twelve task replays and three continuous-process demonstrations. \\
\addlinespace[2pt]
Thermal data interfaces & Prepared & Internal & LeRobot and robomimic formats with small-batch loading checks. \\
\bottomrule
\end{tabularx}
\end{table}

\paragraph{Resources preparation.} The release package is organized as Appendix~\ref{app:release} describes: the simulator core, assets and measured calibration, task definitions, corpus manifests and splits, training and evaluation scripts, and the scripts that regenerate each table and figure, together with the LeRobot and robomimic data interfaces.

\subsection{Integration cost and scope}
\label{subsec:applications-performance}

Table~\ref{tab:integration-cost} records what an integrator can plan for: the measured cost of a coupled payload run and the boundaries of the physical model. A workload with 165,997 thermal nodes over 885 update steps and 115.5~s of simulated time runs in 291.1~s of wall-clock time on a single A6000 node, with the thermal network on the host CPU. The contact-off variant takes 292.4~s, and the six-run cup sweep averages 294.2~s under the same workload dimensions (Table~\ref{tab:runtime-sup}).

\begin{table}[H]
\centering\small
\caption{Integration scope. Measured payload workload, followed by model scope and extension directions. The thermal network runs on the host CPU.}
\label{tab:integration-cost}
\renewcommand{\arraystretch}{1.16}
\begin{tabularx}{\linewidth}{@{}>{\raggedright\arraybackslash}p{0.30\linewidth}>{\raggedright\arraybackslash}X@{}}
\toprule
\multicolumn{2}{@{}l}{\textbf{Measured cost of one coupled run}}\\
\midrule
Thermal nodes & 165,997 \\
Update steps & 885 \\
Simulated time & 115.5~s \\
Coupled payload wall-clock & 291.1~s \\
Contact-off wall-clock & 292.4~s \\
Cup sweep wall-clock & 294.2~s (mean of six runs) \\
Hardware & Single A6000 node; thermal network on the host CPU \\
\midrule
\multicolumn{2}{@{}l}{\textbf{Model scope and extensions}}\\
\midrule
Contact resistance & Fixed and area-normalized per material pair; force- and roughness-dependent correlations are a natural extension (Appendix~\ref{app:scope}). \\
Convection & Fixed convection coefficient; coefficient sweeps are a natural extension. \\
Radiation & Broadband radiative terms; view-factor sweeps are a natural extension. \\
Measured evidence & Scene-specific thermal baseline and camera calibration; aligned real dynamic heat-transfer records are a natural extension. \\
\bottomrule
\end{tabularx}
\end{table}

\subsection{Outlook}
\label{subsec:outlook}

The results point to several directions that follow directly from the engine.

\begin{itemize}
\setlength{\itemsep}{2pt}\setlength{\parskip}{0pt}
\item \textbf{Calibrated dynamic digital twins.} Extend measured initialization and camera calibration to identify dynamic heat-transfer parameters from aligned real and simulated records.
\item \textbf{Heating-dependent task programs.} Generate heating, quenching and reheating histories systematically to study how prior use changes the next decision.
\item \textbf{Thermal data for embodied policy learning.} Extend the decision studies to full robot policies, more objects and materials, and real RGBT observations.
\item \textbf{Force- and roughness-dependent contacts.} Introduce contact-resistance models that respond to normal force and surface roughness, evaluated through the same contact-power interventions.
\item \textbf{Broader platform coverage.} Add robot configurations, backends and data interfaces while retaining the shared workflow.
\end{itemize}

\section{Conclusion}
\label{sec:conclusion}
TRR attaches thermal state to material regions on both contact partners and carries the spatial and internal heat storage that actions leave behind. Contact, reset, transport and depth interventions show how this state changes a later exchange and its safety consequence. A common execution clock connects that evolving state to camera observations, event-local checks and paired RGBT records. A dedicated TRR-generated image-to-site corpus supports safe choices from visible thermal history. Regional decisions with shared feedback execution improve simulated transfer completion with dynamic training data under a new heating procedure. Through reusable assets, two backend adapters, a configurable thermal camera and structured records, the same engine takes a robot scene to thermally labeled data and to executed task evaluation. The paired benchmark, the developer workflow of Figure~\ref{fig:workflow} and the resources of Section~\ref{sec:applications} give external teams a concrete starting point, and the project page brings demonstrations and resource updates together.

\clearpage

{
    \small
    \clearpage
\bibliographystyle{ieeenat_fullname}
    \bibliography{TRR_references}
}

\clearpage
\onecolumn
\raggedbottom
\trrappendixheading
\appendix
\noindent\textbf{Reading guide.}
Appendix~\ref{app:workflow-comparison} compares related workflows, and Appendix~\ref{app:interfaces} details the thermal and camera models. Appendix~\ref{app:experimental-details} gives the experimental protocols: the primary image-to-site study is in Appendix~\ref{app:dynamic-decision-protocol}, and native regional decisions with shared execution are in Appendix~\ref{app:native-modular055}. Additional mechanism checks appear in Appendix~\ref{app:additional-tables}; continuous execution records are collected in Appendix~\ref{app:additional-mechanism-records}. Appendix~\ref{app:release} describes the reproducibility package.

\section{Comparison with Related Workflows}
\label{app:workflow-comparison}

\begin{table}[H]
\trrtablelayout
\centering
\caption{Thermal state, interaction coupling and infrared output in representative workflows.}
\label{tab:comparison-scope}
\begingroup
\footnotesize
\setlength{\tabcolsep}{2pt}
\renewcommand{\arraystretch}{1.18}
\begin{tabularx}{\textwidth}{>{\raggedright\arraybackslash}p{0.195\textwidth}>{\raggedright\arraybackslash}p{0.15\textwidth}>{\raggedright\arraybackslash}X>{\raggedright\arraybackslash}p{0.14\textwidth}>{\raggedright\arraybackslash}p{0.155\textwidth}c}
\toprule
{\bfseries Workflow} & {\bfseries Thermal state} & {\bfseries Interaction coupling} & {\bfseries IR output} & {\bfseries Measured evidence} & {\bfseries \shortstack{Robot\\tasks}} \\
\midrule
iGibson 2.0 \cite{li2021igibson} & Object states & Temperature-dependent state changes & --- & --- & $\checkmark$ \\
BEHAVIOR-1K / OmniGibson \cite{li2023behavior} & Object temperature & Source and ambient exchange & --- & --- & $\checkmark$ \\
HAZARD \cite{zhou2024hazard} & Object temperature & Evolving fire and object temperatures & --- & --- & $\checkmark$ \\
RFUniverse \cite{fu2022rfuniverse} & Particle material state & Contact-dependent material change & --- & --- & $\checkmark$ \\
Tactile Genesis \cite{chung2026tactile} & Link temperature voxels & Sensing voxels coupled to an opposing link scalar & --- & --- & $\checkmark$ \\
\addlinespace[3pt]
Stonefish \cite{grimaldi2502stonefish} & Prescribed surface texture & --- & Thermal-camera image & --- & $\checkmark$ \\
Gazebo thermal camera \cite{gazebothermal} & Prescribed temperature & --- & Thermal-camera image & --- & $\checkmark$ \\
HEARTH \cite{su2026hearth} & Measured apparent-temperature field & Fixed-field manipulation rollouts & Mapped apparent temperature & Measured field & $\checkmark$ \\
\addlinespace[3pt]
DIRSIG \cite{dirsigthermal,rs4082477,brown2010verification} & Surface temperature field & --- & IR sensor model & Calibration, evaluation & --- \\
OKTAL-SE \cite{oktal2022} & Surface temperature field & --- & IR sensor model & Calibration, evaluation & --- \\
ThRend \cite{pedro2021thrend} & External thermal field & --- & IR scene radiance & Evaluation & --- \\
\midrule
\rowcolor{trrmethod}\textbf{TRR} & Material-local temperature & Local heat exchange between both contact partners & Temperature, grayscale & Field, calibration, evaluation & $\checkmark$ \\
\bottomrule
\end{tabularx}
\endgroup
\end{table}

\paragraph{Contact conductance and damage-aware simulators.}
The interface uses a fixed area-normalized contact resistance; force- and roughness-dependent correlations (Yovanovich-style models~\cite{yovanovich2005four}) are outside the current scope and a natural refinement. Damage-aware manipulation simulators such as HAZARD encode temperature as a condition or exposure proxy for safety constraints, whereas TRR supplies the evolving thermal state that such constraints consume.

\section{Model and Calibration Details}
\label{app:interfaces}
The following sections follow the computation from initialization to heat exchange, surface readout and camera output. Surface-field fitting specifies the measured baseline; the final section distinguishes numerical checks from measured validation.

\subsection{RGB-thermal observations across backends}
\label{app:backend-rgbt}
The MuJoCo and Isaac Sim adapters provide native RGB and TRR temperature observations of their respective simulation states (Figure~\ref{fig:backend-rgbt}). These independent examples use a registered fixed thermal field on Piper in MuJoCo and the native-atlas thermal observer on Franka in Isaac Sim.

\subsection{Thermal-state initialization}
\label{app:thermal-calibration}

Task initialization forms $\mathbf T_0$ from an object's reference distribution $\mathbf T_{\mathrm{base}}$:
\begin{equation}
\mathbf T_0=\mathcal U_{\mathrm{task}}(\mathbf T_{\mathrm{base}}).
\label{eq:thermal-initialization}
\end{equation}
The operator applies the requested overall or regional temperature adjustments and leaves the reference unchanged when no adjustment is specified.

Measured baselines use the temperature convention of their calibration data. In the M600 configuration, $\mathbf A$ contains session-aligned apparent temperatures in degrees Celsius. These values supply an effective surface-temperature estimate,
\begin{equation}
T_{\mathrm{base},p}=\mathcal S_p[\mathbf A]+273.15,
\label{eq:measured-baseline}
\end{equation}
in kelvin. Here $\mathcal S_p$ samples the reconstructed field at registered surface location $p$. Objects without measurements use a specified default distribution. The session offset in Eq.~\eqref{eq:m600-session-offset} belongs to camera reporting; calibrated surface thermometry supplies surface-temperature baselines directly.

The task-initialized surface distribution supplies the initial thermal-cell values. Surface-network cells use $T_i(0)=T_{0,i}$. For the layered objects used here, all cells in a depth column start at the temperature assigned to its surface location:
\begin{equation}
T_{p,\ell}(0)=T_{0,p},\qquad \ell=1,\ldots,L_p.
\label{eq:layered-initialization}
\end{equation}
Here $p$ identifies a surface column and $\ell$ its depth layer.

\subsection{Thermal discretization}
\label{app:thermal-discretization}

A thin surface cell uses $V_i=A_id_i$. Layered assets divide the represented material into depth cells. Their masses and heat capacities sum to the asset's represented mass and total heat capacity.

Adjacent finite-volume cells use the shared interface area and the two material resistances:
\begin{equation}
G_{ij}=\frac{A_{ij}}{\ell_i/k_i+\ell_j/k_j},
\label{eq:fv-conductance}
\end{equation}
where $A_{ij}$ is the conducting interface area and $\ell_i,\ell_j$ are the distances from the cell centers to that interface. The layered interface conductance is given in Eq.~\eqref{eq:layered-contact-conductance}.

For the split update, internal conduction and reference-area convection are implicit:
\begin{equation}
\frac{\mathbf C}{\Delta t_h}(\mathbf T^{k+1}-\mathbf T^k)
+(\mathbf L+\mathbf H)\mathbf T^{k+1}
=\mathbf H T_{\mathrm{air}}\mathbf 1+\mathbf Q_{\mathrm{exp}}^k.
\label{eq:implicit-thermal-update}
\end{equation}
Here $\mathbf C$ is the diagonal heat-capacity matrix and $\mathbf L$ is the internal conductance matrix. The diagonal entries $H_{ii}=h_iA_i^{\mathrm{ref}}$ use the fixed boundary area before contact coverage. Let $A_i^{\mathrm{exp},k}$ be the area exposed to the environment during substep $k$. The explicit power contains contact exchange, source power, radiation and a covered-area convection correction:
\begin{equation}
Q_{\mathrm{exp},i}^k
=Q_i^{\mathrm{contact},k}+Q_i^{\mathrm{source},k}
+Q_i^{\mathrm{rad},k}+Q_{\mathrm{cover},i}^k,
\label{eq:split-explicit-power}
\end{equation}
\begin{equation}
\begin{split}
Q_{\mathrm{cover},i}^k
&=h_i\bigl(A_i^{\mathrm{ref}}-A_i^{\mathrm{exp},k}\bigr)
(T_i^k-T_{\mathrm{air}}),\\
Q_i^{\mathrm{rad},k}
&=\varepsilon_i\sigma A_i^{\mathrm{exp},k}
\bigl[T_{\mathrm{air}}^4-(T_i^k)^4\bigr].
\end{split}
\label{eq:split-boundary-power}
\end{equation}
The correction compensates for convection on the contact-covered part of the reference boundary. Contact exchange and radiation use cell temperatures at $t_k$, while the reference-area convection is implicit in $\mathbf T^{k+1}$. Source power is supplied for the current substep. Emissivity $\varepsilon_i$ sets radiative exchange with the ambient environment. The fully implicit contact configuration includes contact conductances in the implicit system. Split configurations choose thermal substeps according to the explicit exchange rates while preserving the elapsed mechanical time.

\paragraph{Contact configurations and patch aggregation.}
\label{app:contact-configurations}
The area-resolved configuration uses separate cell-pair couplings as in Eq.~\eqref{eq:contact-exchange}. Contact activation and effective area follow the geometry and contact model of the workload. Backend force information can gate active contacts; force-dependent patch models additionally use force to estimate a patch extent. Geometry-projected coverage and configured gripper contact areas are other supported choices. These configurations share conservative power exchange but differ in the represented interface.

In the compact patch configuration, each side reads a weighted local temperature:
\begin{equation}
\bar T_a=\sum_i w_i^a T_i,\qquad
\bar T_b=\sum_j w_j^b T_j,\qquad
\sum_iw_i^a=\sum_jw_j^b=1.
\label{eq:contact-patch-state}
\end{equation}
Weights are nonnegative. The patch conductance determines the exchange, and the same weights distribute power back to the participating cells:
\begin{equation}
q_{ab}=G_{ab}^{\mathrm{contact}}(\bar T_b-\bar T_a),
\qquad Q_i^a=w_i^a q_{ab},\quad Q_j^b=-w_j^b q_{ab}.
\label{eq:compact-contact-exchange}
\end{equation}
Normalized weights give equal and opposite total power. A patch selecting one cell on each side reduces to Eq.~\eqref{eq:contact-exchange}; averaging a larger patch retains only its aggregate temperature difference. Multiple couplings accumulate into the corresponding cell contact powers. Spatial detail is consequently determined by the configured cell and patch resolution.

\subsection{Sampling visible surface temperatures}
\label{app:surface-sampling}

After the thermal state is advanced, TRR computes the surface temperature of a layered asset from its current outer-cell temperature and contact heat input. The readout boundary relation accounts for half-cell conduction, convection and ambient radiation:
\begin{equation}
\begin{split}
\frac{T_i^{\mathrm{surf}}-T_i}{r_i}
={}&q_i^{\mathrm{contact}}-f_i\Bigl[h_i(T_i^{\mathrm{surf}}-T_{\mathrm{air}})\\
&+\varepsilon_i\sigma\bigl((T_i^{\mathrm{surf}})^4-T_{\mathrm{air}}^4\bigr)\Bigr].
\end{split}
\label{eq:surface-boundary-balance}
\end{equation}
Here $r_i=d_i/(2k_i)$ is the area-normalized half-cell resistance, $f_i=A_i^{\mathrm{exp}}/A_i$ is the exposed fraction, and $q_i^{\mathrm{contact}}=Q_i^{\mathrm{contact}}/A_i$ is contact power per unit cell surface area. Emissivity $\varepsilon_i$ determines ambient radiative exchange. At readout, contact power is evaluated from the current cell temperatures using Eq.~\eqref{eq:contact-exchange}, with the half-cell resistances included through Eq.~\eqref{eq:layered-contact-conductance}. Given this input and the current $T_i$, TRR solves Eq.~\eqref{eq:surface-boundary-balance} for $T_i^{\mathrm{surf}}$. The convective and radiative terms in this readout relation use the recovered surface temperature. Configured source power enters the cell balance in Eq.~\eqref{eq:thermal-balance}.

Surface-network cells use their state temperature directly. The resulting surface values supply camera observations and surface-temperature task checks. They are averaged at vertices within the same object and region, then interpolated across the visible triangle:
\begin{equation}
\widetilde T_v=\frac{\sum_{i\in\mathcal G(v)}A_iT_i^{\mathrm{surf}}}{\sum_{i\in\mathcal G(v)}A_i},
\qquad T^{\mathrm{surf}}(u)=\sum_{j=1}^{3}\beta_j(u)\widetilde T_{v_j}.
\label{eq:surface-reconstruction}
\end{equation}
Here $\mathcal G(v)$ contains cells sharing vertex $v$, and $\beta_j(u)$ are barycentric weights at surface point $u$. Surface regions preserve specified thermal boundaries. Emissivity and reflection parameters are defined in the material frame and move with the object. Diffuse or rough specular reflection can use directional environment samples or temperatures at visible scene surfaces.

\subsection{Camera response and display mapping}
\label{app:camera-response}

The two observation models supply equivalent signals for common camera processing. The direct-temperature model encodes the current surface temperature in kelvin as
\begin{equation}
\widetilde L_{\mathrm{direct}}(u,t)
=\frac{\sigma[T^{\mathrm{surf}}(u,t)]^4}{\pi}.
\label{eq:direct-temperature-signal}
\end{equation}

For either observation model, spatial filtering and pixel integration operate on the assembled equivalent signal:
\begin{equation}
S^{\mathrm{sp}}_{c,t}=\mathcal P_c\!\left(G_c*\widetilde L_{c,t}\right),
\label{eq:camera-spatial-response}
\end{equation}
where $G_c$ is a normalized Gaussian kernel and $\mathcal P_c$ averages the samples assigned to each pixel. The first-order temporal response updates the camera's retained signal $H_c$:
\begin{equation}
H_{c,t}=H_{c,t^-}+\left(1-e^{-\Delta t_{c,t}/\tau_c}\right)
\left(S^{\mathrm{sp}}_{c,t}-H_{c,t^-}\right),
\qquad S_{c,t}=H_{c,t}+\eta_{c,t}.
\label{eq:temporal-response}
\end{equation}
Here $\tau_c$ is the response time and $\Delta t_{c,t}$ is the elapsed camera observation interval. The first observation initializes $H_c$ from $S^{\mathrm{sp}}_{c,t}$; disabling temporal lag uses the current spatial signal directly. Noise $\eta_{c,t}$ is added after the history update and is not written back into $H_c$. For configured noise-equivalent temperature difference $n_c$, radiance-domain noise has standard deviation $n_c\varepsilon_{\mathrm{ref},c}4\sigma T_{\mathrm{ref},c}^3/\pi$, using the reference temperature and emissivity supplied in the sensor configuration. The positive processed signal is converted to temperature by Eq.~\eqref{eq:radiance-readout}, followed by display mapping.

The optical readout is converted to Celsius for the M600 configuration.

A base display mapping uses fixed temperature limits or image percentiles $q_{\mathrm{lo}}<q_{\mathrm{hi}}$:
\begin{equation}
\mathcal D_{\mathrm{base}}(Y)=255\operatorname{clip}\!\left(
\frac{Y-q_{\mathrm{lo}}}{q_{\mathrm{hi}}-q_{\mathrm{lo}}},0,1\right).
\label{eq:agc}
\end{equation}
The measured display model adds normalized global and local contrast features. For input matrix $Z$, let $\zeta_L(Z)$ be the linear temperature-range mapping, $\zeta_H(Z)$ a global cumulative-distribution mapping, and $\zeta_C(Z)$ a contrast-limited local cumulative-distribution mapping. Their mixture passes through a fitted curve and is blended with the base response:
\begin{align}
\zeta(Z)&=w_L\zeta_L(Z)+w_H\zeta_H(Z)+w_C\zeta_C(Z),
\quad w_r\geq0,\quad\sum_rw_r=1,\notag\\
F(\zeta)&=c_0+\sum_{k=0}^{K-1}c_{k+1}\operatorname{clip}(K\zeta-k,0,1),\notag\\
\mathcal D_{\phi_c}(Z)&=(1-\gamma)\mathcal D_{\mathrm{base}}(Z)+\gamma F(\zeta(Z)).
\label{eq:measured-agc}
\end{align}
The coefficients of $F$ are nonnegative and satisfy $\sum_{k=0}^{K}c_k\leq255$, with $0\leq\gamma\leq1$. Fitting uses $Z=Y$ from measured temperature--grayscale pairs; rendering uses the simulated optical readout $Z=\widehat Y^{\mathrm{opt}}$.

The M600 assets use a reference apparent-temperature field to align capture sessions. For session $s$, its alignment offset is estimated from training correspondences:
\begin{equation}
b_s=\operatorname{median}_{p\in\mathcal P_s^{\mathrm{train}}}
\operatorname{median}_{q\in\Omega_p}
\left[Y_p(q)-A_p^{\mathrm{ref}}(q)\right].
\label{eq:session-alignment}
\end{equation}
Here $\Omega_p$ contains trusted image--surface correspondences, and $A_p^{\mathrm{ref}}$ is the reference asset projected into view $p$. Surface fitting aggregates $Y_p-b_s$. The retained M600 output offset is $b=-0.4490\,^{\circ}$C, estimated from 40 training frames.

In the M600 configuration, measured baselines and task-specific initial temperatures use the same calibrated reference scale. The offset $b$ applies to the complete numerical output after spatial response, while display processing uses the preceding optical readout:
\begin{equation}
\widehat Y_C^{\mathrm{out}}=\widehat Y_C^{\mathrm{opt}}+b,
\qquad \widehat I=\mathcal D_{\phi_c}(\widehat Y_C^{\mathrm{opt}}).
\label{eq:m600-session-offset}
\end{equation}
The M600 display normalizes each image using its temperature quantiles and relative contrast. These features remove a common temperature offset, allowing the same display fit to serve native measured matrices and session-aligned simulated readings. Temperature errors use $\widehat Y^{\mathrm{out}}$, and grayscale errors use $\widehat I$. The ten-configuration evaluation uses this calibration-session alignment. The material-dependent observation configuration uses the unshifted readout in Eq.~\eqref{eq:radiance-readout}.

Camera intrinsics and view registration specify projection. The M600 configuration uses Gaussian widths $\sigma_x=\sigma_y=0.7$ native pixels and three subpixel samples per axis. Temporal response and NETD are device settings; emissivity and reflected surroundings are scene inputs for the material-dependent model.

\subsection{Surface-field fitting}
\label{app:field-fitting}

For surface patches reconstructed by the graph model, the session-aligned reading $y_{pn}$ at node $n$ from pose $p$ constrains the apparent-temperature field $\mathbf A$ and auxiliary local gradients $\mathbf g$. The objective combines a weighted Huber data term, local-gradient regularization and a temperature prior:
\begin{equation}
\begin{split}
\min_{\mathbf A,\mathbf g}\quad
&\sum_{p,n}w_{pn}\rho_\delta(A_n-y_{pn})\\
&+\lambda_c\!\sum_{(i,j)\in\mathcal E_s}e_{ij}
\left[A_j-A_i-\tfrac12(\mathbf g_i+\mathbf g_j)\cdot(\mathbf x_j-\mathbf x_i)\right]^2\\
&+\lambda_g\ell_s^2\!\sum_{(i,j)\in\mathcal E_s}e_{ij}v_{ij}
\|\mathbf g_j-\mathbf g_i\|_2^2
+\sum_n\eta_n^2(A_n-A_n^{\mathrm{prior}})^2.
\end{split}
\label{eq:measured-field-fit}
\end{equation}
Here $\rho_\delta$ is the Huber loss, $w_{pn}$ weights a reading's measurement support, and $A_n^{\mathrm{prior}}$ constrains regions with limited observations. The $\lambda_c$ penalty aligns neighboring temperature differences with their estimated gradients, and the $\lambda_g$ penalty smooths gradients across neighboring nodes. Here $\mathcal E_s$ is surface adjacency, $\mathbf x_n$ is node position, $e_{ij},v_{ij}$ are edge weights, and $\ell_s$ sets the regularization length. Field values, observations and priors use common acquisition units, Celsius in the native workflow.

Session-aligned measurements are aggregated in local tangent planes and fitted by iteratively reweighted sparse least squares. Local corrections retain surrounding asset values and region boundaries. The fitted field supplies the measured baseline through Eq.~\eqref{eq:measured-baseline}; Table~\ref{tab:field-fit-config} gives the fitting settings.

The Piper asset combines reference surface fields, adopted graph fits on detail patches, and two subsequent local refinements. Graph edges follow mesh adjacency within each surface patch, with $e_{ij}=1$ and $v_{ij}=\max(0.1,\max(0,\mathbf n_i\cdot\mathbf n_j)^4)$. Table~\ref{tab:field-fit-config} identifies the settings used for these fits.

Observation weights balance sessions and repeated poses while retaining geometric confidence. Let $c_{pn}$ be the local confidence, $N_s$ the number of selected sessions, $G_s$ the number of pose groups in session $s$, and $N_{sg}$ the number of frames in its pose group $g$. The weights are
\begin{equation}
\begin{split}
u_{pn}&=\frac{c_{pn}}{N_sG_{s(p)}N_{s(p)g(p)}},\\
\bar c_n&=\frac{\sum_pu_{pn}c_{pn}}{\sum_pu_{pn}},
\qquad w_{pn}=\bar c_n\frac{u_{pn}}{\sum_pu_{pn}}.
\end{split}
\label{eq:fit-observation-weights}
\end{equation}
Unsupported or nonfinite observations have zero weight. Nodes supported by fewer than three distinct pose groups have no data term and are constrained by the prior and graph regularization. Within the local neighborhood, sample weights are a Gaussian of width 1.75~mm multiplied by incidence and correspondence confidence. Confidence scales with $\min(N/6,1)$ for $N$ accepted samples; coarse correspondences receive a factor of 0.35. The local incidence factor is $\operatorname{clip}[(|\mathbf n\cdot\mathbf v|-0.2)/0.6,0.2,1]$. Frame--node confidence is its weighted average, including correspondence confidence. For fewer than five local samples, aggregation uses the weighted mean; otherwise it fits a robust tangent-plane model.

\begin{table}[htbp]
\trrtablelayout
\centering\small
\caption{Fitting settings for the measured Piper asset. Graph parameters apply to the adopted detail patches; the two local refinements retain surrounding asset values.}
\label{tab:field-fit-config}
\begin{tabularx}{\textwidth}{@{}>{\raggedright\arraybackslash}p{0.30\textwidth}>{\raggedright\arraybackslash}X@{}}
\toprule
{\bfseries Stage / parameter} & {\bfseries Setting} \\
\midrule
\textbf{Graph-field fitting} & \\
Graph data loss & Huber threshold $0.35\,^{\circ}$C; four reweighting iterations \\
Graph reference length & $\ell_s=3$~mm \\
Graph regularization & Matrix strength $s\in\{0.003,0.01,0.03\}$; energy coefficients in Eq.~\eqref{eq:measured-field-fit} are $\lambda_c=\lambda_g=s^2$ \\
Prior / gradient anchor & $\eta_n=0.003$ with at least three supporting pose groups, otherwise 0.15; gradient anchor matrix strength $10^{-6}$ \\
Surface node spacing & 3~mm on detail surfaces; 4.5~mm on the D435 main housing; reference link-2 main field uses 6~mm \\
Selection and adopted strengths & Five folds grouped by pose; minimize MAE plus half the mean discrepancy of the 10th, 50th and 90th percentiles. Adopted $s$: 0.03 on two link-2 detail faces; 0.003/0.03/0.003 on three bracket faces; 0.003 and 0.01 on the powered and unpowered main housing; 0.003 on another powered D435 face \\
\addlinespace[3pt]
\textbf{Local surface fitting} & \\
Local correspondence support & Radius 3.5~mm, at most 16 samples; normal dot product $\geq0.9$ and absolute incidence cosine $\geq0.3$ \\
Local tangent-plane fit & Huber threshold $0.25\,^{\circ}$C, three iterations; slope ridge $0.02$ times the summed robust sample weights \\
\addlinespace[3pt]
\textbf{Asset refinements} & \\
Joint-3 cap refinement & First-order surface plane; coordinate scale $25\times24$~mm; Huber $0.35\,^{\circ}$C; ridge 0.025; nine training frames \\
D435 housing residual fit & Support radius 10~mm; normal threshold 0.7; Huber $0.5\,^{\circ}$C, six iterations; ridge matrix strength 0.06 (quadratic coefficient 0.0036) \\
\bottomrule
\end{tabularx}
\end{table}

\makeatletter
\setlength{\@fptop}{0pt}
\makeatother
\subsection{Model boundaries and validation scope}
\label{app:scope}
\paragraph{Physical approximations.}
Interface conductance uses a fixed area-normalized resistance per material pair. Normal force can activate contact, but neither force nor roughness changes this resistance. Convection uses a fixed coefficient; radiation and reflected radiance use broadband approximations. The present checks therefore do not constitute an exhaustive sweep of convection coefficients, emissivities or view factors.

\paragraph{Measured observation validation.}
The Piper study evaluates a scene-adaptation workflow: measured thermal fields define an asset baseline, measured temperature--grayscale pairs fit the camera response, and the resulting configuration is fixed for ten scoring poses excluded from fitting. The scoring images come from one M600 capture session. Apparent-temperature and grayscale errors are reported separately in Appendix~\ref{app:m600-pose-reuse}. The mixed manifest also contains 28 real recordings used for reader and integrity checks.

\paragraph{Dynamic validation.}
Analytical contact and diffusion references check numerical accuracy and energy conservation (Appendix~\ref{app:task-variations}); controlled interventions test the consequences of retained thermal state. Measured post-separation cooling traces provide an ordinal check on decay (Appendix~\ref{app:cooling-depth}). The available physical setup differs from the simulated contact scenes in task objects, contact geometry and observation viewpoints. These measurements support thermal-asset calibration, camera-response evaluation and exposed cooling comparisons. Joint identification of dynamic heat-transfer parameters would require aligned real and simulated records of object geometry, contact footprints, boundary conditions and measurement times. Such matched contact records are outside the current acquisition.

\section{Experimental Configurations and Data Protocol}
\label{app:experimental-details}
These studies answer different questions. Mechanism interventions isolate heat exchange and retained state; archive checks assess trajectory labels; image-to-site tests score learned choices; native execution tests score the resulting robot task. Their datasets, thresholds and evaluation units are specified separately. The regional-decision study in Appendix~\ref{app:native-modular055} evaluates executed transfer with a common feedback controller.
\subsection{Interaction configurations}
\label{app:task-variations}
The transfer and cooling runs use 10,503 thermal nodes: a lumped aluminium shell, two layered supports and two finite gripper-pad states. Spatial fields are resolved on the supports; the shell has one temperature state. The ambient and initial shell temperatures are 25 and $70\,^{\circ}$C. The support and pad interface resistances are $0.002$ and $0.005\,\mathrm{m^2K/W}$, respectively, with a $650\,\mathrm{mm^2}$ effective area per active pad. The separate contact-hold run uses a compact patch solver and silicone pad. Contact footprints come from projected object geometry, activate under a positive normal force, accumulate per cell pair and are capped by the receiving-cell area; partial occlusion reduces the exposed area used by convection and radiation.

Stored camera frames match state indices and physical timestamps. The synthetic camera uses a $640\times512$ image, a $43^\circ$ field of view and $3\times3$ subpixel sampling. Broadband graybody emission uses emissivities of 0.90 for the shell and 0.95 for the other rendered surfaces, with ambient reflection at $25\,^{\circ}$C. Recorded grayscale maps 25--$75\,^{\circ}$C to the display range.

For the transfer run, halving the maximum hold step from 0.10 to 0.05~s changes the 40$\,^{\circ}$C trigger time by 0.05~s. Across phase-matched samples, the maximum shell and board-surface differences are $0.011$ and $0.118\,^{\circ}$C, respectively.

\begin{figure}[!t]
\centering\includegraphics[width=\linewidth]{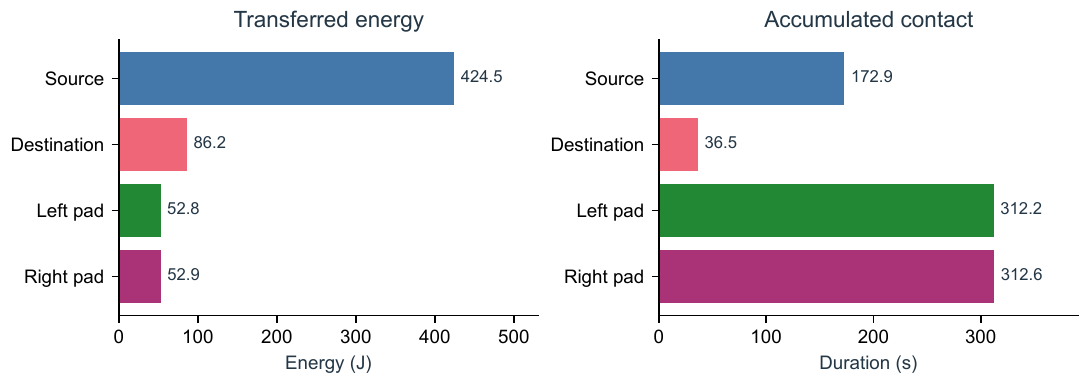}
\caption{Recorded contact-channel totals for the transfer execution. Energy is accumulated from object-to-partner exchange; duration is accumulated separately for each active channel. Concurrent channels can contribute during the same interval.}
\label{fig:contact-channel-totals}
\end{figure}
The transfer exchanges heat with both supports and both gripper pads (Figure~\ref{fig:contact-channel-totals}). Per-channel contact durations sum to 834.15~s within a 519.8~s execution because channels can be active concurrently.

\begin{figure}[!t]
\centering\includegraphics[width=\linewidth]{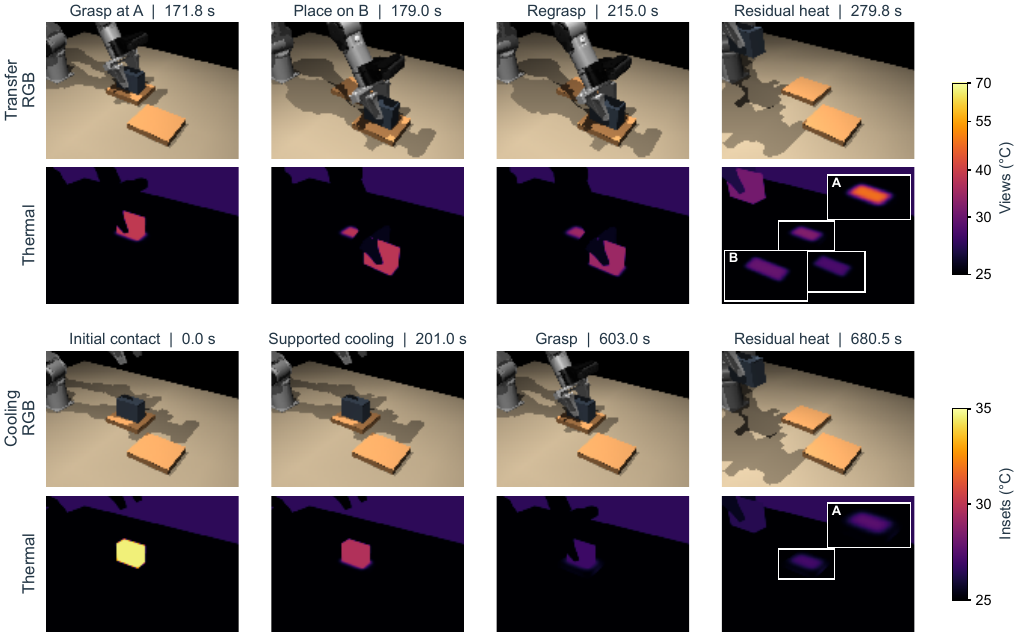}
\caption{Transfer and regrasp (top); cooling and removal (bottom). Full thermal views share a 25--70$\,^{\circ}$C scale; insets show residual heat at A and B on a 25--35$\,^{\circ}$C scale.}
\label{fig:process-checks}
\end{figure}

\paragraph{Camera-triggered transfer.}
\label{app:camera-trigger}
We repeat the same Piper grasp, lift, transfer and release from initial shell temperatures of 60, 70 and $80\,^{\circ}$C. During the fixed-pose wait, the external camera samples apparent temperature at 2~Hz using the projected shell mask, 0.7-pixel spatial blur and independent 0.15$\,^{\circ}$C Gaussian pixel noise. The readout averages pixels inside the rendered shell silhouette after a three-pixel erosion. The radiometric trigger first inverts the graybody relation with the configured emissivity (0.90) and ambient temperature ($25\,^{\circ}$C). Each trigger starts the same motion when its reading falls to $40\,^{\circ}$C. The simulation state supplies the reference trigger and scores physical temperature at the instant the motion starts.

\paragraph{Emissivity sensitivity of the camera trigger.}
The closed-loop runs use the configured emissivity $\epsilon_0=0.90$. Because the controller inverts the graybody relation, an error in $\epsilon$ propagates directly to the physical temperature at which the emissivity-corrected estimate reaches 40$\,^{\circ}$C and triggers motion (Table~\ref{tab:emissivity-sensitivity}). A $\pm0.05$ error shifts the physical trigger by about $0.8\,^{\circ}$C, smaller than the $1.5\,^{\circ}$C gap between the apparent-temperature and corrected triggers reported above.

\begin{table}[H]
\trrtablelayout
\centering\small
\caption{Physical trigger temperature when the emissivity-corrected estimate reaches 40$\,^{\circ}$C with assumed $\epsilon_0=0.90$.}
\label{tab:emissivity-sensitivity}
\begin{tabular}{@{}lrrrrr@{}}
\toprule
{\bfseries True emissivity $\epsilon$} & {\bfseries 0.80} & {\bfseries 0.85} & {\bfseries 0.90} & {\bfseries 0.95} & {\bfseries 1.00} \\
Shell temperature at trigger ($^{\circ}$C) & 41.73 & 40.82 & 40.00 & 39.26 & 38.59 \\
\bottomrule
\end{tabular}
\end{table}

All nine runs mechanically complete the transfer, with a 0.142~m maximum lift and a 3~mm final horizontal error from the target center. The radiometric trigger satisfies the temperature constraint in all three camera-controlled runs, while the three apparent-temperature triggers start above the physical limit (Table~\ref{tab:camera-trigger}).

\begin{table}[!htbp]
\trrtablelayout
\centering\small
\caption{Camera-triggered transfer. Delay is measured relative to the state trigger. Temperature is the physical shell temperature at motion start.}
\label{tab:camera-trigger}
\begin{tabular}{@{}rrrrrr@{}}\toprule
& & \multicolumn{2}{c}{\bfseries Uncorrected camera} & \multicolumn{2}{c}{\bfseries Corrected camera} \\
\cmidrule(lr){3-4}\cmidrule(l){5-6}
{\bfseries \shortstack{Initial shell\\($^{\circ}$C)}} & {\bfseries \shortstack{State-trigger\\time (s)}} & {\bfseries Delay (s)} & {\bfseries \shortstack{Shell at start\\($^{\circ}$C)}} & {\bfseries Delay (s)} & {\bfseries \shortstack{Shell at start\\($^{\circ}$C)}} \\\midrule
60 & 122.8 & -17.3 & 41.49 & +0.2 & 39.98 \\
70 & 168.8 & -18.8 & 41.53 & +0.2 & 39.98 \\
80 & 206.7 & -19.2 & 41.50 & +0.3 & 39.98 \\
\bottomrule
\end{tabular}
\end{table}

\paragraph{Analytical solver checks.}
Table~\ref{tab:analytical-solver-checks} compares the split thermal solver with two insulated analytical references. In the first, two nodes at 70 and $25\,^{\circ}$C have heat capacities of 20 and 40~J/K and share a 3~W/K contact. Their temperature difference decays at rate $G(1/C_1+1/C_2)$, and their capacity-weighted mean stays fixed. We report the maximum error over both nodes and all integration times through 8~s.

In the second, the initial temperature of a 100~mm aluminum slab is $35+10\cos(\pi x/L)$ in degrees Celsius. The mode amplitude decays as $\exp[-\alpha(\pi/L)^2t]$, with $\alpha=k/(\rho c)$, $k=167$~W/(m\,K), $\rho=2700$~kg/m$^3$ and $c=896$~J/(kg\,K). We compare cell averages at 10~s using a common 0.002~s step. Halving the contact step roughly halves the error, and doubling the slab resolution reduces the error by a factor of 3.7--3.9. The absolute energy residual stays below $4.7\times10^{-11}$~J across both checks.

\begin{table}[!htbp]
\trrtablelayout
\centering\small
\caption{Maximum temperature error against insulated analytical references.}
\label{tab:analytical-solver-checks}
\begin{tabular}{@{}llr@{}}\toprule
{\bfseries Reference} & {\bfseries Resolution} & {\bfseries Max. error ($^{\circ}$C)} \\\midrule
Contact exchange & $\Delta t=0.20$~s & 0.2531 \\
Contact exchange & $\Delta t=0.10$~s & 0.1253 \\
Contact exchange & $\Delta t=0.05$~s & 0.0624 \\
\addlinespace[3pt]
Slab diffusion & 8 cells & 0.0434 \\
Slab diffusion & 16 cells & 0.0112 \\
Slab diffusion & 32 cells & 0.0030 \\
\bottomrule
\end{tabular}
\end{table}

\subsection{Controlled mechanism interventions}
\label{app:mechanism-protocol}
The experiment in Section~\ref{subsec:mechanism-interventions} reuses the transfer example's 10,503-node thermal network and engineering parameters. It isolates a horizontal shell--source interface of $0.0032\,\mathrm{m^2}$ with the same area quadrature and interface resistance. The second support and two pad states remain at ambient and have no contact edges. Contact is active during 0--30~s and 60--90~s, with identical geometry in both intervals. Integration uses 0.1~s steps to 120~s. This is a thermal-only prescribed-contact experiment.

The no-exchange intervention sets contact conductance to zero while retaining geometrical coverage and the same exposed-area boundary conditions. The reset intervention replaces all source-support depth-cell temperatures with $25\,^{\circ}$C once at 30~s; it leaves the object and the subsequent schedule unchanged. This reset removes 260.48~J of stored energy. Accounting for that intervention, the absolute energy-balance residual is below $3.1\times10^{-11}$~J in all variants. Recontact power is evaluated from the current cell temperatures and active conductances immediately before integration at 60~s. The reported precontact surface peak uses the uncovered boundary readout.

The contact-location experiment uses the same plate discretization and a $24\times24$~mm footprint centered at material $x=-30$ or $+30$~mm, with $y=0$. Initial shell and plate temperatures are $70\,^{\circ}$C and $25\,^{\circ}$C. Contact lasts 0--20~s, followed by separation to 60~s. Surface means are queried over the fixed L/R footprints using the uncovered boundary readout. Only footprint position changes between runs. The colder-patch query compares these two means. The scalar readout assigns the arithmetic mean over the uniform plate-surface grid to both patches. The frozen readout assigns the initial plate temperature. Both fields use a common 25--40$\,^{\circ}$C visualization scale.

In the rigid-transport experiment (Figure~\ref{fig:transport-history}), a $120\times100$~mm layered plate starts at $25\,^{\circ}$C. A $70\,^{\circ}$C pad heats a $24\times24$~mm patch during 0--20~s. The plate follows a prescribed rigid transform during 20--26~s and then contacts a second pad during 40--50~s. Both pads have heat capacity 30~J/K; the receiver starts at $25\,^{\circ}$C. We use $R''=0.002\,\mathrm{m^2K/W}$ at each interface. The world-space receiver center maps to the left material patch at $0^\circ$ and to the right at $180^\circ$. The scalar plate replaces its cells by one temperature with their summed capacity, exposed area and parallel contact conductances. Halving the step from 0.10 to 0.05~s changes the initial warm-patch heat flow by 0.003~W.

\subsection{Full static/dynamic corpus}
\label{app:corpus}

\begin{figure}[!t]
\centering
\includegraphics[width=0.58\linewidth]{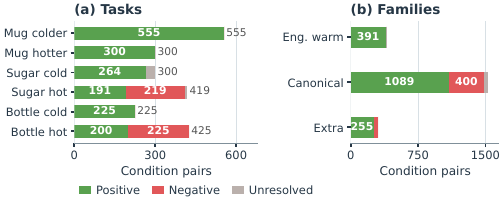}
\caption{Benchmark composition by task and generation family. Bars show dynamic task labels; counts are source-condition pairs.}
\label{fig:paired-coverage}
\end{figure}

Figure~\ref{fig:paired-coverage} groups the 2,224 source-condition pairs into three families: 1,525 original canonical conditions (Canonical), 299 original conditions with completed dynamic counterparts (Extra), and 400 added hot-region conditions (Eng.\ warm). The last family spans 107 source groups with native-initial-state pairing checks. Motion-provenance groups track source motions reused across conditions.

The archive spans 391 source-motion groups. All 444 dynamic violations occur in the original families. The 400 added hot-region conditions use separate thermal initializations and yield 391 passes and nine unresolved outcomes.

The simulated corpus stores 447,318 dual-view observation instants and 110,145,746 native action intervals. The mixed reader manifest additionally includes 28 real recordings and one weak-label task copy.

Record-integrity checks cover pixel-readback and process-exit dependencies, metadata and observation clocks. The four-worker simulation reader makes 192 queries on 130 sources; the mixed reader makes 192 on 121 sources, with a union of 155. These are sampled access checks over the admitted manifests.
The two continuous runs pass frame/state alignment and finite-array checks on all 705 saved camera samples.

\subsection{Contact labels and event timing}
\label{app:contact-labels}
Contact checks use temperature intervals $[T^-,T^+]$ reconstructed from the simulated surface at the task-relevant locations and native times. For a cold-region threshold $\theta_c$, a contact passes when $T^+\leq\theta_c+10^{-5}\,^{\circ}$C and fails when $T^->\theta_c+10^{-5}\,^{\circ}$C. For a hot-region threshold $\theta_h$, it passes when $T^-\geq\theta_h-10^{-5}\,^{\circ}$C and fails when $T^+<\theta_h-10^{-5}\,^{\circ}$C. Intervals crossing the threshold, or locations outside saved reconstruction support, are unresolved. A thermal execution fails if any contact fails, passes if every required contact passes, and is otherwise unresolved; mechanical success is also required for a positive task label.

The certified contact margin is $\theta_c+10^{-5}-T^+$ for cold tasks and $T^--(\theta_h-10^{-5})$ for hot tasks, in degrees Celsius. Figure~\ref{fig:contact-validity} uses the smallest supported margin in each condition and measures first violation delay from the earliest recorded contact. A label-consistency audit covers all 1,369 dynamic records from the four region tasks. Recomputing 61,067,478 threshold checks from saved FP64 temperature intervals across 328 matrix groups reproduces the stored contact statuses and task labels.

\paragraph{Contact coverage.}
We merge qualifying finger--target contacts by distinct 500~Hz step across both fingers. A repeated contact has at least two blocks of 25 or more contact steps separated by at least 100~ms without a qualifying finger--target contact. The counts are 108/300 for sugar-box cold-region, 197/419 for sugar-box hot-region, and 0/225 and 0/425 for bottle cold- and hot-region. The counts are unchanged at 250 and 500~ms gaps. Of the 305 repeated-contact conditions, 108 have a certified violation, and all first violations precede the second contact block. This tally uses the dynamic member of each region-task pair. The other 855 conditions use mug-identity checks. Both benchmark branches share a target, so a rate of history-driven target switching requires a separate choice intervention.

\subsection{Image-based forecast of delayed contact violations}
\label{app:forecast}
We generate 576 planar contact sequences in 96 groups, with six preheating variants per group. A group fixes contact position, receiver resistance and initial temperature, and the camera projection, texture and emissivity. The variants change initial plate temperature, heating-pad temperature, dwell and cooling intervals. Labels use the mean physical surface temperature of the contacted patch after the receiver touches it. The forecast set retains the 555 trials whose first contact meets $34\,^{\circ}$C. A positive forecast label marks a crossing below this threshold within 1.5~s at a 0.05~s solver step. Groups are split into 72 training groups (418 trials, 126 crossings) and 24 test groups (137 trials, 36 crossings).

The orthographic camera samples $32\times32$ RGB and apparent-temperature crops around the upcoming contact region. Two images are sampled at $-0.60$ and $-0.10$~s relative to contact. The temperature channel uses graybody emission, ambient reflection, a 0.7-pixel Gaussian blur and 0.15$\,^{\circ}$C Gaussian noise. Frozen inputs use the initial plate field and the same camera settings and noise as the dynamic inputs. The logistic probe uses $4\times4$ average pooling per channel and frame, training-set standardization and the same $L_2$ regularization. RGB-only uses zero-filled thermal channels. Test AUROC is 0.517, 0.626 and 0.978 for RGB, frozen-field RGBT and evolving-field RGBT, respectively.

Test crossings occur 0.05--1.50~s after touch (median 0.675~s). At a probability threshold of 0.5, the early-frame and two-frame probes each detect all 15 crossings within 0.50~s and 17 of the 21 later crossings.

\paragraph{Matched histories before a common pinch.}
\label{app:matched-history}
Each of 48 groups starts with a $120\times100$~mm three-layer coupon in a fixture, at a uniform 24--$28\,^{\circ}$C. A heated pad (57--$76\,^{\circ}$C) touches either the future grasp patch or its mirror across the coupon center for 5.50--10.45~s. The coupon then cools for 1.00--3.20~s. Both branches share these settings and a prescribed two-jaw closure with the same poses, camera and contact footprint. MuJoCo confirms both jaw contacts at closure. TRR integrates the upper $20\times20$~mm interface with a 0.05~s step and tracks its patch-mean temperature for 1.5~s. The task marks the first reading above $34\,^{\circ}$C as a violation. Median initial-contact temperatures are 37.37 and $25.81\,^{\circ}$C for same-patch and opposite-patch heating.

\begin{figure}[!htbp]
\centering
\includegraphics[width=0.69\linewidth]{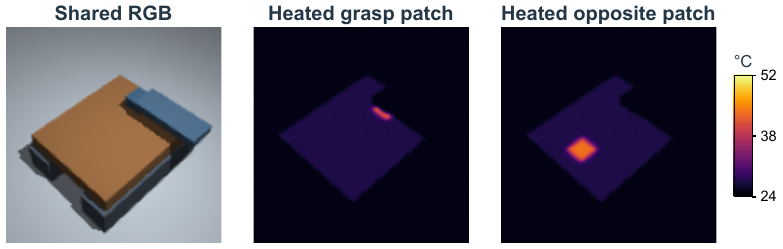}
\caption{Paired observations before the common pinch. Both thermal images use the same apparent-temperature scale.}
\label{fig:matched-history-camera}
\end{figure}

Figure~\ref{fig:matched-history-camera} shows the common precontact view. Depth and segmentation from the rendered view locate visible coupon pixels. The camera samples the surface field 0.20~s before touch, applies graybody emission with emissivity 0.95 and ambient reflection, and adds 0.7-pixel blur and 0.12$\,^{\circ}$C Gaussian noise. Frozen-field images use the initial coupon field with the same projection and noise. The model receives a $32\times32$ crop centered on the prescribed grasp point, pooled to an $8\times8$ grid per channel. Six held-out folds keep both histories of each group together. RGB and frozen-field inputs match exactly within each pair, so each ties all 47 label-changing pairs. The evolving thermal readout orders all 47 correctly.

\paragraph{Recorded thermal observations select a simulated contact site.}
\label{app:recorded-site-selection}
We mark two $5\times5$ patches on the same visible sugar-box face in five pre-approach HIK frames. Their median apparent temperatures differ by 3.25--$9.01\,^{\circ}$C, with unchanged ordering under independent $\pm2$-pixel center shifts along each axis. Three instructions select the cooler patch and two select the warmer patch.

We initialize opposite ends of the fixed three-layer coupon from the apparent temperatures, with a linear transition along its length and the same initial field at every depth. The two MuJoCo closures are 70~mm apart and use the same $20\times20$~mm interfaces. Table~\ref{tab:recorded-site-contrast} reports the simulated contrast in patch-mean upper-layer temperature across branches. Freezing the initialized field preserves the initial contrast and its ordering. A spatially uniform scalar state at the same initial mean ties the symmetric sites. With TRR, the ordering is unchanged and the contrast decreases by 0.51--$1.42\,^{\circ}$C from initialization to the end of the 1.5~s hold.

\begin{table}[!htbp]
\trrtablelayout
\centering\small
\caption{Warm-minus-cool contrast across simulated closures ($^{\circ}$C). Changes are computed relative to initialization before rounding.}
\label{tab:recorded-site-contrast}
\begin{tabular}{@{}clrrrr@{}}
\toprule
{\bfseries Pair} & {\bfseries Target region} & {\bfseries \shortstack{Initial /\\frozen field}} & {\bfseries At contact} & {\bfseries After hold} & {\bfseries Change from initial} \\
\midrule
1 & Cooler & 9.01 & 8.85 & 7.58 & $-1.42$ \\
2 & Cooler & 7.47 & 7.33 & 6.29 & $-1.18$ \\
3 & Cooler & 6.76 & 6.64 & 5.69 & $-1.07$ \\
4 & Warmer & 3.85 & 3.79 & 3.25 & $-0.61$ \\
5 & Warmer & 3.25 & 3.20 & 2.74 & $-0.51$ \\
\bottomrule
\end{tabular}
\end{table}

\subsection{Precontact images from the robot archive}
\label{app:archive-forecast}
Sugar-box hot-region passes and failures cluster by generation family (Figure~\ref{fig:paired-coverage}). We compare violation timing within the original canonical family. Its 200 replays all have a certified violation after initial contact. The target is a first certified violation within 0.25~s of first contact (127 early, 73 later). The native contact clock selects the last original paired camera sample in the 0.10--0.80~s precontact window. Static and dynamic members share the selected native index and time. The observed lead times range from 0.172 to 0.526~s, with a median of 0.314~s. The model receives the images; the native clock supplies sample alignment and labels.

The external and wrist RGB and numerical thermal frames are pooled over their full views on an $8\times8$ grid. Invalid thermal pixels use the median of valid pixels in that image. Each variant shares the same RGB features and uses zero-filled, frozen-field or evolving-field temperature features. Five stratified folds hold out source-motion groups, with standardization and the same $L_2$-regularized logistic model fitted within each training fold. Pooled held-out AP is 0.959 for RGB and 0.993 for both RGBT variants. AUROC is 0.925 for RGB and 0.987 for both RGBT variants. The evolving-field AUROC gain over RGB is 0.061, with a 95\% interval of 0.026--0.110 from resampling the source-motion groups.

\subsection{Native regional decisions with shared feedback execution}
\label{app:native-modular055}
This study evaluates the task utility of simulated thermal training data with mechanical execution shared across arms. The coupon study evaluates contact-site choices; this study evaluates their consequences during robot execution. The task is bottle cold-region transfer. A learned infrared classifier chooses site~0 or site~1 once, before motion. Its action dispatches the corresponding original geometric primitive, whose feedback controller supplies subsequent joint targets from the native state. No task oracle corrects the choice. The controller's legacy names ``cold'' and ``hot'' identify geometric primitives, not their actual temperatures or the correct answer. Heating can reverse which primitive is safe.

\paragraph{Training worlds and actual future-contact labels.}
The fixed source poses are 710000--710002 for training and 710003--710004 for development. Each pose has two reached dynamic histories and one unheated static-precontact world, with both candidate actions actually simulated: 30 valid native branches in total. Mechanical execution and object identity must pass for both candidates; thermal task outcomes must be identified, and at least one candidate must pass. If both pass, the fixed training label is site~0. Labels come from the original future-contact task checks, not a teacher's geometric key or a history identifier. All 30 branches pass this data-eligibility gate.

S uses unheated target fields at $26\,^{\circ}$C, the original common fields on other bodies and a finite wrist at $26.5\,^{\circ}$C. Future contact uses D's finite-capacity bilateral solver. Both S candidates pass; the fixed tie rule labels site~0.

D uses the reached dynamic infrared sequence and its actual dynamic future-contact label. S uses its own unheated observations and actual unheated-world label. C uses D's label but repeats the latest student-visible frame, index~2, into all four slots. The fourth raw physical frame is unavailable to students; D/S instead contain a synthetic $26\,^{\circ}$C occlusion in that slot. Each arm has six training rows and four development rows. S's paired repetitions do not create independent thermal sources. There are only three training poses and two training heat-history templates; this is not a history-family split.

\paragraph{Matched learner and fixed budget.}
The observed float32 Celsius arrays have shape $4\times512\times640$. Rows 220:510 and columns 210:460 are cropped and reduced to $64\times64$ with area interpolation, then encoded as $(T-26)/5$. There is no per-image fitting, clipping or eight-bit quantization. The network flattens the ordered four-frame input, applies a 24-unit tanh layer and a two-logit head. It has 393,290 parameters in four tensors, without BatchNorm or dropout. Equal learner seeds 0, 1 and 2 give identical initial parameters across D/S/C.

All nine fits use the full six-row training batch for exactly 1,000 AdamW updates, learning rate $10^{-3}$, weight decay $10^{-4}$ and gradient clipping at one. A common integer image shift in $[-3,3]^2$ is applied to all rows and frames at each update. The 49 shifts follow one fixed lexicographic cycle repeated to 1,000 updates; encoded-zero padding means $26\,^{\circ}$C. All arms and seeds use the same schedule. This observation nuisance was fixed before the first scientific optimizer step; it adds neither physical histories nor training decisions. Inference has no shift. Only update~1,000 is evaluated, with no checkpoint or hyperparameter selection by development or formal outcomes. Actual optimizer tensors and executed shift schedules are independently checked.

\paragraph{Auxiliary state-readout diagnostic.}
R is a geometric temperature diagnostic, not a static-training baseline. It uses the full latest visible frame and public candidate geometry; learned arms use the fixed crop without geometry masks. Visible target pixels are partitioned by nearest candidate anchor, rather than executed contact footprint. R selects the lower regional maximum apparent temperature. Ties and empty regions select site~0; empty regions also record an abstention without excluding the condition.

\paragraph{Frozen native evaluation and physical branch sharing.}
The same-heating-procedure scenes are 640000--640007; the new-heating-procedure scenes are 650000--650007. Each has histories 0 and 1, giving 16 conditions and eight pose clusters per track. The second track withholds a heating--quenching--opposite-reheating program. Geometry, material and physical law remain unchanged; this is not object-held-out transfer or an unseen constitutive law. Different pose cohorts mean a cross-track difference alone does not isolate the program effect.

At test D and S receive exactly the same dynamic observation, and C repeats its latest visible frame into all four slots. All 288 learned choices and 32 R choices are sealed before any selected formal action. Identical choices at the same complete physical initial state use one actual native execution. The resulting union contains 48 physical branches, 24 per track. This is not 320 independent simulations. Every prediction retains its checkpoint, condition, chosen site and unique physical-branch reference; the representative row does not change the physical action. Before acting, the native wrapper independently verifies the complete MuJoCo integration state, finite thermal field, controller, scene, source closure, time configuration and original criteria. Planning must leave the full integration state unchanged.

Native integration uses a 0.002~s step and at most 60,000 steps. The frozen stop rule retains original failures and completes execution 0.4~s after the feedback controller finishes, including the original final 0.3~s release checks. There is no thermal-oracle choice fallback. The physical gates retain at least 15~mm lift, 0.1~s continuous bilateral airborne hold, 80~mm held transport, final finger forces below 0.02~N, table support fraction at least 0.8, goal overlap, and the original unexpected-arm-collision rule at 0.5~N. Correct-object and all qualifying-contact checks are unchanged. The cold/hot limits remain 28.5/40.5\,$^{\circ}$C, comparison tolerance $10^{-5}\,^{\circ}$C and surface-solve stopping scale $10^{-10}\,^{\circ}$C. These are separate from the $34\,^{\circ}$C coupon/depth mechanism tests. Low-level feedback is closed loop; the high-level thermal decision is single-shot, not infrared visual servoing or wholly learned joint control.

\paragraph{Unknown outcomes and paired interpretation.}
All selected branches remain in the fixed accounting. A certified negative contact point is sufficient to fail the all-points thermal predicate even if other points are unresolved. Without a negative witness, unresolved required points leave the thermal result unknown. A known mechanical or object-identity failure still determines task failure when thermal evidence is absent. We retain these component distinctions. Unresolved task outcomes yield identification bounds on success fractions, expressing outcome uncertainty separately from sampling uncertainty.

For a fixed track, the paired task difference is $16^{-1}\sum_i(Y_{D,i}-Y_{S,i})$. If both policies select the same complete physical branch, that summand is zero even when its absolute outcome is unknown. Thus unresolved outcomes are shared variables; subtracting independent marginal bounds would discard known pairing. Results are reported per learner seed and as a mean across initializations, without counting shared branches or paired histories as independent physical repetitions. The selected-action union supports paired task-outcome comparisons; evaluating preferred-site accuracy would additionally require counterfactual outcomes for every candidate. Development qualifications characterize executor behavior separately from learned-policy evaluation.

\begin{table*}[t]
\trrtablelayout
\centering\footnotesize
\caption{Complete transfer results by heating procedure and learner initialization. D: dynamic training; S: static training; C: latest-visible-frame repeat; R: geometry--temperature rule. Each row has 16 planned conditions in eight pose clusters. Success bounds span certified completions through completions plus unresolved runs. The final columns count mechanical and correct-object passes. A planning failure acts on no object; its false object predicate is not evidence of choosing a distractor.}
\label{tab:native-modular055-full}
\setlength{\tabcolsep}{5pt}
\begin{tabular}{lcrrrcrr}
\toprule
 & & \multicolumn{3}{c}{\bfseries Task outcomes (count)} & & \multicolumn{2}{c}{\bfseries Component passes} \\
\cmidrule(lr){3-5}\cmidrule(l){7-8}
{\bfseries Input} & {\bfseries Seed} & {\bfseries Completed} & {\bfseries Failed} & {\bfseries Unresolved} & {\bfseries \shortstack{Success bounds\\(\%)}} & {\bfseries Mechanical} & {\bfseries \shortstack{Correct\\object}} \\
\midrule
\multicolumn{8}{@{}l}{\textbf{Same heating procedure; held-out placements}} \\
\rowcolor{trrmethod}D & 0 & 12 & 2 & 2 & $[75.00, 87.50]$ & 14/16 & 14/16 \\
\rowcolor{trrmethod}D & 1 & 12 & 2 & 2 & $[75.00, 87.50]$ & 14/16 & 14/16 \\
\rowcolor{trrmethod}D & 2 & 12 & 2 & 2 & $[75.00, 87.50]$ & 14/16 & 14/16 \\
\addlinespace[2pt]
S & 0 & 6 & 9 & 1 & $[37.50, 43.75]$ & 14/16 & 14/16 \\
S & 1 & 6 & 9 & 1 & $[37.50, 43.75]$ & 14/16 & 14/16 \\
S & 2 & 6 & 9 & 1 & $[37.50, 43.75]$ & 14/16 & 14/16 \\
\addlinespace[2pt]
C & 0 & 12 & 2 & 2 & $[75.00, 87.50]$ & 14/16 & 14/16 \\
C & 1 & 12 & 2 & 2 & $[75.00, 87.50]$ & 14/16 & 14/16 \\
C & 2 & 12 & 2 & 2 & $[75.00, 87.50]$ & 14/16 & 14/16 \\
\addlinespace[2pt]
R & -- & 12 & 2 & 2 & $[75.00, 87.50]$ & 14/16 & 14/16 \\
\midrule
\multicolumn{8}{@{}l}{\textbf{New heating procedure; held-out placements}} \\
\rowcolor{trrmethod}D & 0 & 13 & 1 & 2 & $[81.25, 93.75]$ & 15/16 & 15/16 \\
\rowcolor{trrmethod}D & 1 & 13 & 1 & 2 & $[81.25, 93.75]$ & 15/16 & 15/16 \\
\rowcolor{trrmethod}D & 2 & 13 & 1 & 2 & $[81.25, 93.75]$ & 15/16 & 15/16 \\
\addlinespace[2pt]
S & 0 & 5 & 9 & 2 & $[31.25, 43.75]$ & 14/16 & 14/16 \\
S & 1 & 5 & 9 & 2 & $[31.25, 43.75]$ & 14/16 & 14/16 \\
S & 2 & 5 & 9 & 2 & $[31.25, 43.75]$ & 14/16 & 14/16 \\
\addlinespace[2pt]
C & 0 & 13 & 1 & 2 & $[81.25, 93.75]$ & 15/16 & 15/16 \\
C & 1 & 13 & 1 & 2 & $[81.25, 93.75]$ & 15/16 & 15/16 \\
C & 2 & 11 & 3 & 2 & $[68.75, 81.25]$ & 15/16 & 15/16 \\
\addlinespace[2pt]
R & -- & 13 & 1 & 2 & $[81.25, 93.75]$ & 15/16 & 15/16 \\
\bottomrule
\end{tabular}
\end{table*}

\paragraph{Completed outcomes and failure attribution.}

Under the same heating procedure, D has 12 certified successes, two failures and two unresolved outcomes; S has six, nine and one, respectively, out of 16 per initialization (Table~\ref{tab:native-modular055-full}). Six confirmed wins, no confirmed losses, eight known ties, one shared-action unresolved pair and one different-action unresolved pair give a D--S difference of $[37.5,43.75]$ percentage points. In the different-action pair D is unresolved and S fails, giving a contribution of zero or one.

In same-procedure scene640006/history0, D's site~1 action stops after 3,291 steps at 6.582~s: link6 contacts the bottle with normal force 10.734~N, exceeding the unchanged 0.5~N permanent-collision gate. The original native source writes this witness and exits with code~12 before that interval's thermal update. Mechanical failure determines task failure; neither thermal safety nor full-task correct-object success is inferred from the partial grasp. The raw witness and complete terminal mechanical arrays are retained. This branch contributes a mechanical failure to the task results.

Site~0 is unreachable in both histories of same-procedure scene640007 and new-procedure scene650002. Each records the original planning-failure event at time zero, without changing the primitive or falling back to another action. D selects that site in one history per track; S selects it in both. In addition, S has seven certified thermal failures per track. In same-procedure scene640000/history0, the corresponding D action remains unresolved, so only six of these seven thermal contrasts are confirmed wins.

Under the new heating procedure, D has 13 certified successes, one failure and two unresolved outcomes; S has five, nine and two, respectively, per 16 conditions. The eight confirmed wins comprise seven avoided thermal violations and one difference in primitive reachability. There are no losses, six known ties and two shared-action unresolved pairs: the paired difference is exactly 50 percentage points despite unidentified absolute rates.

The two tracks together use 48 actual physical branches: 25 certified successes, 19 failures and four unresolved outcomes. This union accounting is not any individual policy's task rate. D--C is zero on the same-procedure track and zero, zero and 12.5 percentage points across new-procedure learner seeds 0, 1 and 2; D--R is zero on both tracks.

\paragraph{Unresolved physical branches.}

The following complete native branches retain unidentified task outcomes; they are not removed from any policy denominator. The point counts are positive/negative/unresolved under the original contact checker.

\noindent primary\_\_seed640000\_\_h0, site 1: 5559/0/1 points.\par

\noindent primary\_\_seed640002\_\_h1, site 0: 5365/0/77 points.\par

\noindent interaction\_composition\_\_seed650000\_\_h0, site 0: 5412/0/53 points.\par

\noindent interaction\_composition\_\_seed650005\_\_h0, site 0: 5421/0/49 points.\par

\paragraph{Preselected executed example.}
Figure~\ref{fig:native-modular055} uses the first lexicographic seed-0 D/S disagreement in the complete sealed prediction list: interaction-composition scene650000/history1. The example is fixed before observing any formal outcome. Both policies act on the same dynamic observation; D selects site~1 and S site~0. Both complete the original mechanical and correct-object checks. D has 5,888 positive contact-point intervals and no negative or unresolved ones. S has 3,043 positive, 1,781 negative and 641 unresolved intervals. Its negative witnesses determine task failure without discarding the unresolved intervals.

The plotted envelope at each native endpoint is $[\max_j\ell_j,\max_j u_j]$, where $[\ell_j,u_j]$ bounds the simulated temperature at qualifying point $j$. The two maxima need not come from the same point. An upper crossing alone is not a certified negative witness. Raw interval index $k$ refers to the post-integration endpoint; plotting times are checked against the actual native clock at $k+1$, equal to $0.002(k+1)$~s. Missing-contact intervals break the curves; there is no smoothing, temperature clipping or interpolation over gaps. The displayed infrared array is a camera observation, whereas the curves are bounds on simulated contact temperatures. Neither is a new physical contact measurement. The figure uses actual selected branches, never forced qualifications or cached teacher-contact curves.

\subsection{Measured thermal assets across robot configurations}
\label{app:m600-pose-reuse}

\begin{figure}[H]
\centering
\includegraphics[width=0.64\linewidth]{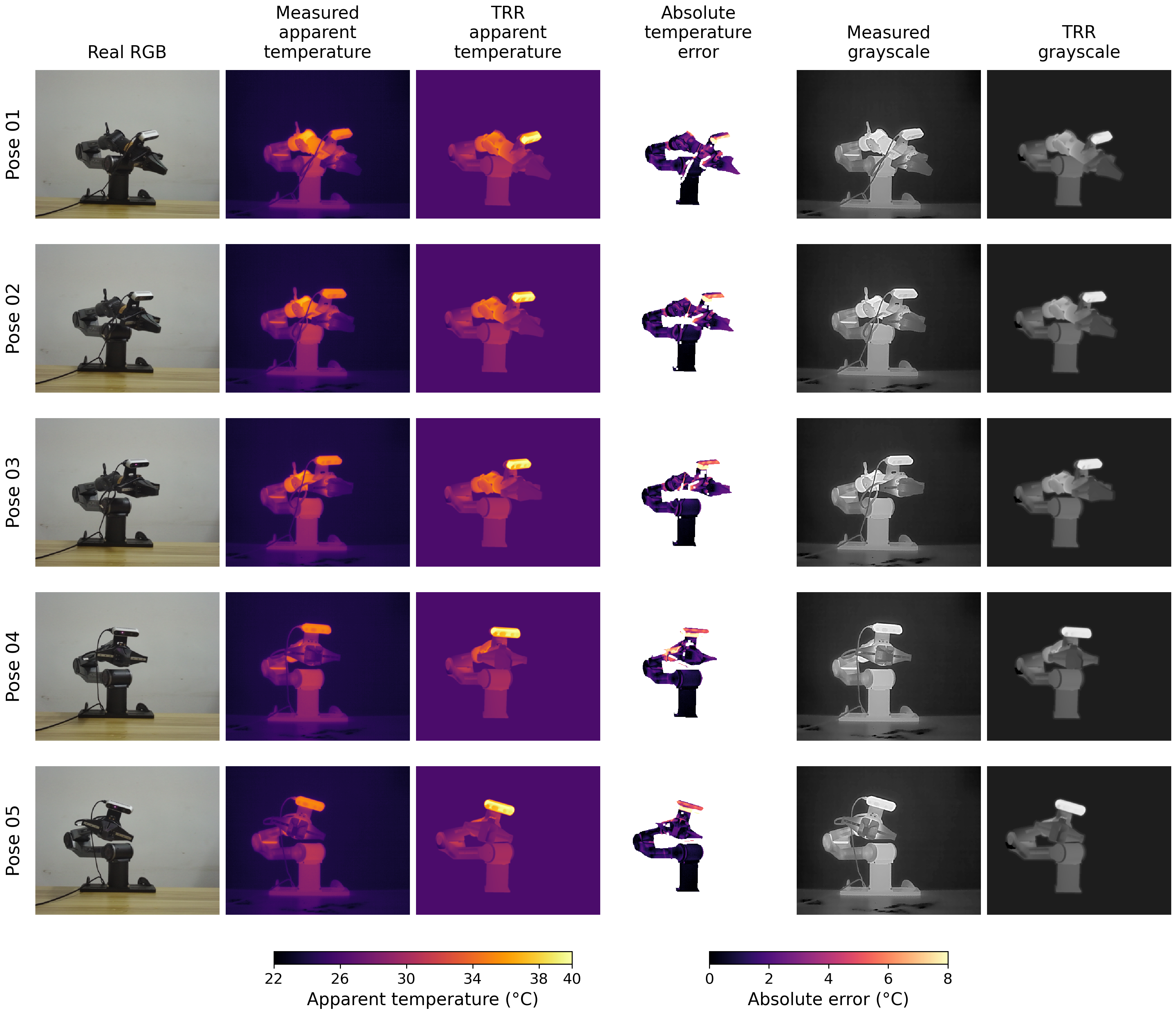}
\caption{Measured and TRR observations for Piper poses 1--5. Shared scales: temperature $22$--$40\,^{\circ}$C, absolute error $0$--$8\,^{\circ}$C, grayscale 0--255. White error-map regions lie outside the evaluation masks.}
\label{fig:m600-all-poses-a}
\end{figure}

\begin{figure}[H]
\centering
\includegraphics[width=0.64\linewidth]{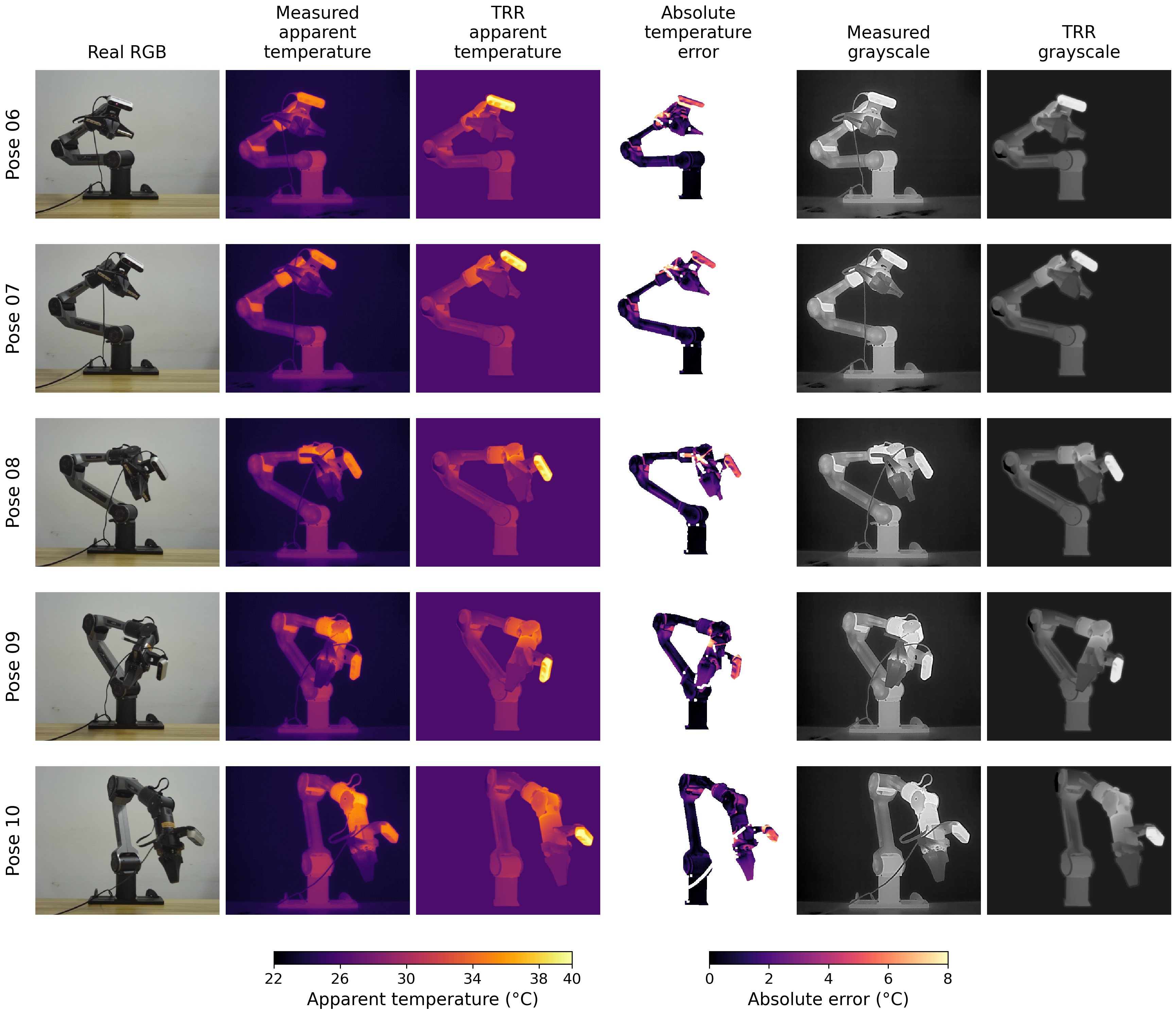}
\caption{Measured and TRR observations for Piper configurations 6--10, with columns and scales as in Figure~\ref{fig:m600-all-poses-a}. All quantitative results: Table~\ref{tab:m600-pose-errors}.}
\label{fig:m600-all-poses-b}
\end{figure}

The comparison uses ten Piper configurations recorded in one M600 capture session under steady operating conditions. The simulated views use the common measured steady-state field and calibrated camera model at the corresponding joint configurations. The camera provides an apparent-temperature matrix and a native grayscale image at $640\times512$ pixels. Figures~\ref{fig:m600-all-poses-a} and~\ref{fig:m600-all-poses-b} show every configuration, together with absolute temperature-error maps. The camera and thermal asset are fixed for scoring, with RGB-based geometry alignment for the session. Figure~\ref{fig:measured-observations} shows pose 7 from the subset chosen before scoring to cover different arm arrangements and occlusions.

Errors are computed on a common evaluation region for temperature and grayscale. A measured foreground mask is obtained from grayscale segmentation within a manually reviewed robot region. This mask and the simulated robot foreground are each eroded by two pixels. The primary mask intersects these eroded foregrounds, excludes unsupported geometry and requires predicted subpixel foreground coverage of at least 0.99. Coverage is the number of evaluated pixels divided by the measured foreground area before erosion. Across the ten configurations, the evaluation contains 178,778 pixel instances and covers 55.03--68.68\% of the measured foreground in each image. Table~\ref{tab:m600-pose-errors} reports each configuration and their equally weighted mean.

Temperature errors compare the reported matrix $\widehat Y^{\mathrm{out}}$ in Eq.~\eqref{eq:m600-session-offset} with the camera-reported apparent temperatures. Grayscale errors use the display image generated from $\widehat Y^{\mathrm{opt}}$. The mean temperature MAE is $1.442\pm0.103\,^{\circ}$C (range $1.279$--$1.583$), and the mean grayscale MAE is $52.15$ gray levels. The mean grayscale bias is $-47.32$ levels. Scores use native image arrays and retain values beyond the plotted ranges.

\begin{table}[!tp]
\trrtablelayout
\centering\small
\caption{Observation errors across ten Piper configurations. Bias is prediction minus measurement; means weight configurations equally.}
\label{tab:m600-pose-errors}
\begingroup\setlength{\tabcolsep}{3pt}
\begin{tabular}{@{}lrrrr@{\hspace{10pt}}rrr@{}}
\toprule
 & & \multicolumn{3}{c}{\bfseries Apparent temperature ($^{\circ}$C)} & \multicolumn{3}{c}{\bfseries Display grayscale (0--255)} \\
\cmidrule(lr){3-5}\cmidrule(l){6-8}
{\bfseries Pose} & {\bfseries \shortstack{Mask coverage\\(\%)}} & {\bfseries MAE} & {\bfseries RMSE} & {\bfseries Bias} & {\bfseries MAE} & {\bfseries RMSE} & {\bfseries Bias} \\
\midrule
1 & 58.88 & 1.528 & 2.179 & +0.480 & 45.76 & 52.91 & -40.77 \\
2 & 58.08 & 1.546 & 2.270 & +0.682 & 48.15 & 54.97 & -42.27 \\
3 & 56.54 & 1.475 & 2.213 & +0.713 & 50.73 & 57.11 & -44.71 \\
4 & 55.03 & 1.583 & 2.508 & +0.938 & 59.30 & 65.64 & -51.91 \\
5 & 55.54 & 1.279 & 2.099 & +0.802 & 58.74 & 64.65 & -54.13 \\
6 & 60.49 & 1.463 & 2.230 & +0.885 & 56.69 & 63.64 & -51.78 \\
7 & 62.20 & 1.446 & 2.092 & +0.935 & 54.39 & 61.23 & -50.65 \\
8 & 64.02 & 1.441 & 2.096 & +0.913 & 50.99 & 58.20 & -46.20 \\
9 & 65.63 & 1.371 & 1.945 & +0.472 & 50.12 & 57.20 & -46.75 \\
10 & 68.68 & 1.289 & 1.820 & +0.098 & 46.64 & 53.68 & -44.03 \\
\midrule
\textbf{Mean} & 60.51 & 1.442 & 2.145 & +0.692 & 52.15 & 58.92 & -47.32 \\
\bottomrule
\end{tabular}\endgroup
\end{table}

\subsection{Measured cooling traces and initialization depth}
\label{app:cooling-depth}
Two residual heat traces from one water-cup transfer are used. The first exposed M600 thermal frame initializes each trace; 13 frames for the first trace and 11 for the second (24 pooled) are scored on the region-mean apparent temperature. The nominal initialization uses 2~mm layers; the table varies this depth for both dynamic models. The frozen baseline keeps the initial field fixed.

\begin{table}[H]
\trrtablelayout
\centering\small
\caption{Pooled region-mean MAE ($^{\circ}$C) against 24 measured cooling frames. Bold marks the lowest error at each initialization depth.}
\label{tab:depth-sensitivity}
\begin{tabular}{@{}lrrr@{}}
\toprule
{\bfseries Initialization depth} & {\bfseries Layered TRR} & {\bfseries Scalar cooling} & {\bfseries Frozen field} \\
\midrule
1~mm & 1.33 & \textbf{1.30} & 2.05 \\
2~mm (nominal) & 0.97 & \textbf{0.93} & 2.05 \\
4~mm & 0.52 & \textbf{0.45} & 2.05 \\
\bottomrule
\end{tabular}
\end{table}

Both dynamic models are closer to the measured decay than the frozen field, and the advantage holds at every depth. The scalar model is slightly better on the region mean, while TRR also retains the spatial pattern of the trace; the nominal 2~mm depth was fixed before this comparison, so the table is a sensitivity check rather than a tuned result. At the 1~mm initialization, the per-trace values are $0.96$ versus $0.93\,^{\circ}$C for the first trace and $1.76$ versus $1.74\,^{\circ}$C for the second; weighting the 13 and 11 frames gives the pooled $1.33$ and $1.30\,^{\circ}$C entries. The nominal 2~mm row uses the pre-registered initialization and is not reconstructed from these 1~mm per-trace numbers. Contact-period heat transfer is not scored here; this check covers the separation-cooling segment that the measured frames resolve.

\subsection{Dynamic decision benchmark}
\label{app:dynamic-decision-protocol}
The primary image-policy comparison is followed by compact-feature studies of source-group partitions, material and object transfer, training-set size and observation perturbations. Each study specifies its source data and evaluation cohort.
The primary evaluation uses 128 test decisions from 64 source groups, excluding validation decisions. The original release protocol additionally appended 16 validation decisions from eight occluded groups; those are retained only as a historical diagnostic and never enter the primary rates. Training uses 256 dynamic source episodes. The independent finite-difference reference certifies the contact outcome; the image policy is an image-to-site readout, not a full-arm robot policy.

Figure~\ref{fig:round10-case} uses the first lexicographic pure-test group in the occluded-contact family; its choices use learner seed~0. The two rows are alternative heating histories, not successive stages of one execution. The observation clock reaches zero at the decision, when a noncontact shutter hides the target; the dark final thermal crop therefore indicates occlusion rather than cooling. Contact begins 0.18~s later. The response plots use a separate clock starting at contact and show contact-surface temperature rather than camera temperature.

\subsubsection{Primary image-to-site comparison}
\paragraph{Model and training.}
The full arm uses a ResNet-18 frame encoder, a GRU over four RGBT frames, and a six-way head for site and wait. The matched final-frame-repeat arm has the same encoder, GRU and head but repeats the decision frame at all four inputs; frozen input and per-frame spatial mean are matched observation substitutions; RGB-only changes input channels; the global-mean GRU removes the spatial encoder. Training is 40 epochs, batch 16, AdamW $10^{-4}$/$10^{-2}$, with a pre-construction RNG reset and one final-epoch evaluation for fixed seeds 0--4. No validation or test selection is performed.

\begin{table}[H]
\trrtablelayout
\centering\small
\caption{Image-to-site test on 128 held-out decisions (64 groups), by learner seed. All rates are percentages. Safe uses the independent finite-difference certificate; site match is exact expert-site agreement. The full and final-frame-repeat arms share the ResNet/GRU/head.}
\label{tab:dynamic-primary-sup}
\begin{tabular}{@{}lrrrr@{}}
\toprule
 & \multicolumn{2}{c}{\bfseries Full RGBT sequence} & \multicolumn{2}{c}{\bfseries Final-frame repeat} \\
\cmidrule(lr){2-3}\cmidrule(l){4-5}
{\bfseries Seed} & {\bfseries Safe (\%)} & {\bfseries Site match (\%)} & {\bfseries Safe (\%)} & {\bfseries Site match (\%)} \\
\midrule
0 & 95.31 & 92.19 & 75.00 & 62.50 \\
1 & 94.53 & 82.03 & 75.00 & 62.50 \\
2 & 92.97 & 79.69 & 75.00 & 62.50 \\
3 & 88.28 & 75.00 & 75.00 & 63.28 \\
4 & 87.50 & 75.00 & 75.00 & 62.50 \\
\midrule
Mean & \textbf{91.72} & \textbf{80.78} & 75.00 & 62.66 \\
\bottomrule
\end{tabular}
\end{table}

The observed full-minus-final-frame-repeat difference is 16.72 points, with source-group bootstrap interval [11.56,22.03]. Frozen input has the same observed 75.00/62.66. The fixed third-frame crop rule reaches 85.16/74.22; it is a public geometric rule, not a learned policy. The full model is 6.56 points above it, with group interval [3.43,10.31], and the case in the main text shows that the rule also succeeds on the selected pair.

\paragraph{Group bootstrap.}
For each source group, we first average the two histories and the five fixed seeds. We then pair the two arms at the group level, resample the 64 pure-test groups with replacement 2000 times, and report the percentile 95\% interval with random seed 20261110. The 640 model-case records are five seeds times 128 decisions; they are not 640 independent tests. No per-seed or per-history record is treated as an independent group.

\subsubsection{Channel and frame-only controls}
\label{app:channel-study}
\paragraph{Matched channel comparison.}
This channel study trains its own full-RGBT baseline, so comparisons use the rows within this block rather than the primary baseline. The table reports test rates after excluding validation decisions.

\begin{table}[H]
\trrtablelayout
\centering\small
\caption{Matched channel study, held-out test (128 decisions; seeds 0--4). Bold and underline mark the highest and second-highest observed percentages. The legacy last-frame path differs in BN/input handling; the corrected frame-only five-seed control is reported separately.}
\label{tab:dynamic-channel-sup}
\begin{tabular}{lrr}
\toprule
{\bfseries Input / model} & {\bfseries Safe (\%)} & {\bfseries Site match (\%)} \\
\midrule
\rowcolor{trrmethod}Full RGBT sequence & \underline{90.00} & \underline{77.34} \\
Thermal only & \textbf{99.06} & \textbf{99.06} \\
RGB only & 73.91 & 61.41 \\
Frozen input & 75.00 & 62.50 \\
Per-frame spatial mean & 73.13 & 60.63 \\
Uniform global mean & 71.88 & 59.38 \\
Legacy final-frame model & 75.00 & 70.94 \\
\bottomrule
\end{tabular}
\end{table}

Full minus thermal-only is $-9.06$ points (group interval $[-13.29,-5.31]$); the channel result is retained and does not support a claim that RGB is necessary. The corrected frame-only control is 75.0\% safe, site match 75.0/73.4/74.2/74.2/65.6 over seeds 0--4; it is a strategy diagnostic, not a matched GRU control.

\subsubsection{Stability across source-group partitions}
\paragraph{Source-group re-partitioning.}
Each of five source-group partitions trains a separate compact-feature policy and evaluates 128 test decisions from 64 groups. Its 512-entry representation contains a 128-value final-frame vector and three frame-difference blocks of 128 values. The evaluation sets share 11--24 groups pairwise. Within each split, confidence intervals use resampling of the 64 source groups.

\begin{table}[H]
\trrtablelayout
\centering\small
\caption{Compact-feature policy across five source-group partitions. Each test contains 128 decisions from 64 groups; the mean weights the five fits equally.}
\label{tab:dynamic-split-sup}\label{tab:dynamic-splits}
\begin{tabular}{@{}lrrrr@{}}
\toprule
{\bfseries Split} & {\bfseries \shortstack{Sequence\\safe (\%)}} & {\bfseries \shortstack{Final-frame\\safe (\%)}} & {\bfseries \shortstack{Paired gain\\(points)}} & {\bfseries \shortstack{95\% group-bootstrap\\interval (points)}} \\
\midrule
s0 & 92.97 & 75.00 & +17.97 & [11.72,24.22] \\
s1 & 90.63 & 75.00 & +15.63 & [10.16,21.09] \\
s2 & 92.19 & 75.00 & +17.19 & [11.72,22.66] \\
s3 & 90.63 & 75.00 & +15.63 & [10.16,21.09] \\
s4 & 91.41 & 75.00 & +16.41 & [10.16,22.66] \\
\midrule
Mean & 91.56 & 75.00 & +16.56 & — \\
\bottomrule
\end{tabular}
\end{table}

\subsubsection{Transfer across material configurations}
\paragraph{Material-held-out evaluation.}
A second corpus uses three material sets on the same rectangular coupon geometry (300 groups / 1200 episodes; train M1, validation M3, test M2 held out by material). The four-frame sequence reaches 99.5\% site accuracy on the material-held-out test histories versus 50.0\% for last-frame; the metadata last-visible rule reaches 100\% using privileged information. This split isolates material variation at fixed geometry; the object-held-out test below changes the geometry while retaining the contact-history task family.

\begin{table}[H]
\trrtablelayout
\centering\small
\caption{Material-held-out corpus (300 groups / 1200 episodes). The coupon geometry is unchanged; material set M1 is used for training, M3 for validation and M2 for testing. Bold and underline mark the highest and second-highest accuracies (\%).}
\label{tab:material-generalization-sup}
\begin{tabular}{lrrr}
\toprule
{\bfseries Input} & {\bfseries \shortstack{Test accuracy\\(M2)}} & {\bfseries \shortstack{Balanced test\\accuracy (M2)}} & {\bfseries \shortstack{Validation accuracy\\(M3)}} \\
\midrule
\rowcolor{trrmethod}4-frame sequence & \underline{99.5} & \underline{99.5} & \underline{99.0} \\
Last frame & 50.0 & 50.0 & 50.0 \\
Last visible frame (privileged metadata) & \textbf{100.0} & \textbf{100.0} & \textbf{100.0} \\
\bottomrule
\end{tabular}
\end{table}

\subsubsection{Source-trained shared-candidate transfer across objects}
\label{app:shared-candidate-transfer}
Three thermal-feature policies are fitted on the coupon source data: dynamic-sequence D, static-sequence S, and final-frame L. Training uses 256 matched source episodes. D and L use the dynamic source labels; S uses the static source observations and labels. Source-fitted feature scales and coefficients are fixed for target evaluation.

The target assets are a mug and a sugar box with 86,674 and 48,391 layered thermal nodes. Each supplies 32 groups with two mirrored histories, giving 64 decisions per object and 128 overall. Each group has a uniform initial temperature drawn from 24.8--25.5\,$^{\circ}$C, a 6~s contact with a finite 24~J/K heating pad initially at 90\,$^{\circ}$C, followed by 4~s of separation. The pads are externally insulated; exposed object surfaces use $h=8\,\mathrm{W\,m^{-2}K^{-1}}$ and $\epsilon=0.95$, with contact resistance $0.0082166\,\mathrm{m^2K/W}$. Observations are saved at 1, 2, 3 and 4~s after separation; a noncontact shutter hides the final view. Contact regions depend only on geometry. A simplified orthographic surface-cell projection generates target images, and each candidate's $32\times32$ thermal crop supplies a mean temperature at each observation time. 

Both site choices are continued from the common decision state using a finite 9~J/K pad and native TRR contact updates. A safe choice keeps the area-weighted conducting contact-patch surface below 34\,$^{\circ}$C for 1.5~s. The peak includes the initial contact readout and 30 subsequent samples at 0.05~s intervals. TRR rollout certificates score each site choice using these contact-surface continuations. The native thermal graphs, contact resistance, source-model parameters, code and seed ranges are frozen before target evaluation. An auxiliary independent finite-volume check on four fixed graph histories with a 60\,$^{\circ}$C pad (25--46.6\,$^{\circ}$C fields) is a numerical implementation check.

\paragraph{Shared-candidate source readout.}
For site $i$, we use four public crop features
\[
\phi_i=[\mu_{i,4},\;\mu_{i,4}-\mu_{i,1},\;\mu_{i,4}-\mu_{i,2},\;\mu_{i,4}-\mu_{i,3}],
\]
where $\mu_{i,t}$ is the mean of the site's $32\times32$ thermal crop. After source-fitted component scaling $\sigma$, the shared linear score gives class-1 logit
\[
\ell=\theta^\top[(\phi_0-\phi_1)\oslash\sigma],
\]
with component-wise division. Positive logits select site~1 and exact zero selects site~0. Each arm fits source-only scaling and coefficients once. D uses dynamic source observations and dynamic labels; S uses matched static observations and their original static labels; L uses dynamic labels with the three temporal differences zeroed during both fitting and evaluation. Source standardization scales without centering and logistic regression fits no intercept; the common configuration is $C=1$, LBFGS, maximum 2000 iterations and tolerance $10^{-8}$.

The target evaluation uses seeds 330000--330031 for mug and 340000--340031 for sugar, with 32 groups and 64 decisions per object. It retains the geometry-defined sites, physical contact protocol, camera/shutter, 34\,$^{\circ}$C surface criterion and native continuation. Inference extracts public crop means from archived thermal arrays using the frozen source scaler and coefficients. The source-fitted policies are evaluated on this registered target cohort using the same thermal task and contact continuation for all three arms. The dynamic source validation manifest contains 64 rows from 32 groups, with 48, 40 and 41 correct choices for D, S and L, respectively.

On this cohort, D, S and L yield 128/128, 0/128 and 64/128 safe selections; per object, D/S/L are 64/64, 0/64 and 32/64. D selects site~1 on h0 and site~0 on h1 for both objects. S selects the unsafe opposite site in both histories. L has exact-zero logits on all 128 decisions because the common shutter makes the fourth-frame difference zero, and the zero tie selects site~0. The paired D--S outcomes are D-only 128, both 0; D--L is both 64 and D-only 64. The cohort retains all 128 decisions; generation failures, invalid predictions, no-safe and multi-safe cases are all zero.

\begin{table}[H]
\trrtablelayout
\centering\small
\caption{Shared-candidate transfer from coupon source training to two target objects. Each object supplies 32 groups and 64 paired-history decisions. Cells report safe choices; exact agreement has the same counts.}
\label{tab:shared-candidate-transfer-sup}
\begin{tabular}{@{}lrrrr@{}}
\toprule
{\bfseries Target object} & {\bfseries Dynamic (D)} & {\bfseries Static (S)} & {\bfseries Final frame (L)} & {\bfseries \shortstack{D--S gain\\(points)}} \\
\midrule
Mug & $64/64$ & $0/64$ & $32/64$ & $+100.0$ \\
Sugar box & $64/64$ & $0/64$ & $32/64$ & $+100.0$ \\
\textbf{Overall} & $128/128$ & $0/128$ & $64/128$ & $+100.0$ \\
\bottomrule
\end{tabular}
\end{table}

\subsubsection{Training-set scaling and observation perturbations}
\label{app:scaling-perturbation}
The compact-feature readout is also evaluated as the training set grows and as observation noise increases. The scaling study uses the trend and occluded-contact families, increasing the fraction of training groups from 10\% to 100\%. Safe choices rise from 71.3\% to 87.5\%. The observation-perturbation study uses the archived feature-policy evaluation of 128 test and 16 validation decisions: thermal noise of 0.5 and 1.0\,$^{\circ}$C gives 91.7\% and 87.5\% safe choices, compared with 94.4\% under nominal observations. The recorded RGB-noise and blur settings each retain 94.4\%.

\begin{table}[H]
\trrtablelayout
\centering\small
\caption{Training-set scaling and observation perturbations for the compact-feature policy. Training fractions use the trend and occluded families; perturbations use the archived 144-decision evaluation. Each block retains all its tested settings.}
\label{tab:scaling-perturbation-sup}
\begin{tabular}{@{}llr@{}}
\toprule
{\bfseries Study} & {\bfseries Setting} & {\bfseries Safe choices (\%)} \\
\midrule
Training data fraction & $10\%$ & 71.3 \\
& $25\%$ & 83.3 \\
& $50\%$ & 86.3 \\
& $100\%$ & 87.5 \\
\midrule
Observation perturbation & Nominal & 94.4 \\
& Thermal noise $0.5\,^{\circ}$C & 91.7 \\
& Thermal noise $1.0\,^{\circ}$C & 87.5 \\
& RGB noise & 94.4 \\
& RGB blur & 94.4 \\
\bottomrule
\end{tabular}
\end{table}

\section{Additional Tables}
\label{app:additional-tables}
The following tables supplement the mechanism and forecast studies. The payload example compares local temperature peaks with a body-mean scalar, then varies the contact path and material parameters. The final subsection gives the hidden-depth counterexample protocol used in the main paper.

\begin{table}[H]
\trrtablelayout
\centering\small
\caption{Plate-mean scalar baseline for the safety decision. The scalar is fitted to the same cup--phone--support traces; it predicts the payload mean, whereas the decision uses the localized peak.}
\label{tab:scalar-baseline-sup}
\setlength{\tabcolsep}{4pt}
\begin{tabular}{@{}lrrrr@{}}
\toprule
{\bfseries \shortstack{Initial vessel\\($^{\circ}$C)}} & {\bfseries \shortstack{Peak body mean\\($^{\circ}$C)}} & {\bfseries \shortstack{Local peak\\($^{\circ}$C)}} & {\bfseries Scalar: action / outcome} & {\bfseries TRR: action / outcome} \\
\midrule
$40\,^{\circ}$C & 25.14 & 27.57 & place / safe & place / safe \\
$50\,^{\circ}$C & 25.23 & 29.26 & place / safe & place / safe \\
$60\,^{\circ}$C & 25.31 & 30.93 & place / safe & place / safe \\
$70\,^{\circ}$C & 25.39 & 32.58 & place / safe & place / safe \\
$80\,^{\circ}$C & 25.47 & 34.22 & place / too hot & avoid / safe \\
$90\,^{\circ}$C & 25.56 & 35.85 & place / too hot & avoid / safe \\
\bottomrule
\end{tabular}
\end{table}

\begin{table}[H]
\trrtablelayout
\centering\small
\caption{Payload workload and measured wall-clock. Timing is single-process on one A6000 node; the thermal network is CPU-side split-explicit.}
\label{tab:runtime-sup}
\setlength{\tabcolsep}{4pt}
\begin{tabular}{@{}lrrrr@{}}
\toprule
{\bfseries Run} & {\bfseries Thermal nodes} & {\bfseries Update steps} & {\bfseries \shortstack{Simulated\\time (s)}} & {\bfseries \shortstack{Elapsed\\time (s)}} \\
\midrule
Payload, coupled & 165,997 & 885 & 115.5 & 291.1 \\
Payload, contact-off & 165,997 & 885 & 115.5 & 292.4 \\
Cup sweep, mean (6 runs) & 165,997 & 885 & 115.5 & 294.2 \\
\bottomrule
\end{tabular}
\end{table}

\paragraph{Direct versus support path.}
The payload hold has five vessel--phone links and one phone--desk support link; there is no vessel--desk link. Removing the phone--desk links leaves the peak at 34.95\,$^{\circ}$C, while removing the vessel--phone links leaves it at 25.02\,$^{\circ}$C. The all-links peak is 35.85\,$^{\circ}$C.

\begin{table}[H]
\trrtablelayout
\centering\small
\caption{Path ablation for the 90\,$^{\circ}$C payload hold. Only the indicated body-pair contact links are removed.}
\label{tab:path-ablation-sup}
\begin{tabular}{lrr}
\toprule
{\bfseries Active contact links} & {\bfseries Peak body mean ($^{\circ}$C)} & {\bfseries Local peak ($^{\circ}$C)} \\
\midrule
\rowcolor{trrmethod}All links (coupled) & 25.555 & 35.846 \\
Vessel--phone only & 25.569 & 34.945 \\
Phone--desk only & 25.008 & 25.016 \\
No contact heat & 25.000 & 25.000 \\
\bottomrule
\end{tabular}
\end{table}

\paragraph{Bounded material sensitivity.}
We vary one factor at a time around the configured values: contact resistance $0.001$--$0.004$~m$^2$K/W, conductivity $0.5$--$2\times$, heat capacity $0.7$--$1.5\times$, and emissivity $0.8$--$1.0$. At 90\,$^{\circ}$C every variant still produces an over-temperature peak (34.17--35.63\,$^{\circ}$C), while the plate-mean scalar remains 25.36--25.84\,$^{\circ}$C and therefore still places the vessel. The decision margin is at least 1.17\,$^{\circ}$C; this is a one-factor bounded check, not a full factorial or a calibrated uncertainty model.

\begin{table}[H]
\trrtablelayout
\centering\small
\caption{Bounded material/contact sensitivity at the 90\,$^{\circ}$C decision. Each row changes one factor from the configured baseline (max peak 35.85\,$^{\circ}$C); all spatial decisions remain over-temperature.}
\label{tab:material-sensitivity-sup}
\setlength{\tabcolsep}{4pt}
\begin{tabular}{@{}lrrr@{}}
\toprule
{\bfseries Factor} & {\bfseries Tested range} & {\bfseries \shortstack{Peak body mean\\($^{\circ}$C)}} & {\bfseries \shortstack{Local peak\\($^{\circ}$C)}} \\
\midrule
Contact resistance (m$^2$K/W) & 0.001--0.004 & 25.56--25.58 & 34.83--35.14 \\
Conductivity & $0.5$--$2\times$ & 25.36--25.84 & 34.17--34.84 \\
Heat capacity & $0.7$--$1.5\times$ & 25.48--25.63 & 34.21--35.63 \\
Emissivity & $0.8$--$1.0$ & 25.56--25.59 & 34.87--35.25 \\
\bottomrule
\end{tabular}
\end{table}

\paragraph{Same-mean history probe.}
Ten matched histories place the same 90\,$^{\circ}$C vessel on one half of the phone at offsets $\pm0.006$ to $\pm0.018$~m. Within each mirrored pair the phone mean peaks differ by at most 0.083\,$^{\circ}$C, while the warmer/cooler half reverses; the spatial contrast (0.44--0.87\,$^{\circ}$C) is an order of magnitude larger than the mean difference. The task is to contact the cooler half. The full spatial state selects it in all ten histories; a leave-one-out scalar threshold on the phone mean reaches 2/10, and a reset field or fixed-side policy reaches 5/10. The ten-case probe isolates the representation difference that the six-row contact-off table cannot show.

\begin{table}[H]
\trrtablelayout
\centering\small
\caption{Matched-mean history probe. Contact offset is along the phone's $y$ axis; positive offsets heat the $y_+$ half. Decisions select the cooler half.}
\label{tab:same-mean-history-sup}
\setlength{\tabcolsep}{4pt}
\begin{tabular}{@{}lrrrrr@{}}
\toprule
{\bfseries Offset (m)} & {\bfseries \shortstack{Body mean\\($^{\circ}$C)}} & {\bfseries \shortstack{$y_-$ half mean\\($^{\circ}$C)}} & {\bfseries \shortstack{$y_+$ half mean\\($^{\circ}$C)}} & {\bfseries \shortstack{Local maximum\\($^{\circ}$C)}} & {\bfseries Cooler half} \\
\midrule
$-0.018$ & 25.504 & 25.770 & 25.238 & 39.163 & $y_+$ \\
$-0.016$ & 25.499 & 25.757 & 25.240 & 38.916 & $y_+$ \\
$-0.012$ & 25.541 & 25.846 & 25.235 & 35.181 & $y_+$ \\
$-0.008$ & 25.582 & 25.885 & 25.280 & 34.866 & $y_+$ \\
$-0.006$ & 25.571 & 25.803 & 25.338 & 35.014 & $y_+$ \\
\addlinespace[3pt]
$+0.006$ & 25.535 & 25.260 & 25.809 & 34.592 & $y_-$ \\
$+0.008$ & 25.500 & 25.246 & 25.753 & 38.201 & $y_-$ \\
$+0.012$ & 25.459 & 25.238 & 25.680 & 35.126 & $y_-$ \\
$+0.016$ & 25.463 & 25.030 & 25.896 & 39.100 & $y_-$ \\
$+0.018$ & 25.435 & 25.035 & 25.836 & 41.661 & $y_-$ \\
\bottomrule
\end{tabular}
\end{table}

\begin{table}[H]
\trrtablelayout
\centering\small
\caption{Decision arms on the ten matched histories. Scalar uses only the phone mean with a leave-one-out threshold; reset removes the half pattern after separation; full TRR reads the two half means.}
\label{tab:same-mean-arms-sup}
\begin{tabular}{lrl}
\toprule
{\bfseries State representation} & {\bfseries Correct / total} & {\bfseries Decision input} \\
\midrule
Plate-mean scalar & 2/10 & Phone mean only \\
Reset field & \underline{5/10} & Uniform field after separation \\
\rowcolor{trrmethod}Full TRR & \textbf{10/10} & Two half means \\
\bottomrule
\end{tabular}
\end{table}

An in-sample threshold on the same scalar means reaches 8/10, but its leave-one-out accuracy falls to 2/10; the spatial state requires no threshold fitting across histories.

\begin{table}[H]
\trrtablelayout
\centering\small
\caption{Precontact forecast by modality and frame time. Average precision (AP) intervals use source-group bootstrap resampling; Bold and underline mark the two highest distinct point estimates per metric.}
\label{tab:forecast-timing-sup}
\setlength{\tabcolsep}{3pt}
\begin{tabular}{@{}lrrr@{}}
\toprule
{\bfseries Input} & {\bfseries Average precision} & {\bfseries 95\% interval} & {\bfseries Recall} \\
\midrule
RGB & 0.261 & 0.188--0.367 & 0.056 \\
Frozen-field RGBT & 0.337 & 0.245--0.493 & 0.333 \\
Dynamic, early thermal & \underline{0.941} & 0.842--0.995 & \textbf{0.889} \\
Dynamic, late thermal & \textbf{0.942} & 0.843--0.995 & \underline{0.861} \\
Dynamic, both thermal & 0.939 & 0.838--0.995 & \textbf{0.889} \\
\bottomrule
\end{tabular}
\end{table}

\subsection{Hidden-depth counterexample protocol}
\label{app:hidden-depth-protocol}
Two identical slabs use a $1$~cm$^2$ face, $L=6$~mm thickness, $k=0.2$~W/mK and $\rho c_p=1500\times1000$~J/m$^3$K, starting at $25\,^{\circ}$C in a $25\,^{\circ}$C environment. The model is one-dimensional through thickness with finite volumes; side and bottom faces are adiabatic and radiation is disabled. A surface heater is $80\,^{\circ}$C with $h=2000$~W/m$^2$K for $0.05$~s or $180$~s. A controlled quench is $25\,^{\circ}$C with $h=5000$~W/m$^2$K; the two quench durations are $0.0125719586$~s and $3.5119399975$~s. After the quench the heater/cooler is removed, both slabs release for $0.1$~s under natural $h=8$~W/m$^2$K, and only then is the decision read. The future contact is a finite pad at $25\,^{\circ}$C with $C=0.5$~J/K, the same $1$~cm$^2$ area, $R''=0.0082166$~m$^2$K/W, pad convection $h=8$~W/m$^2$K, and $1.5$~s duration; the object contacts only through this pad. The reported peak is the object-side reconstructed interface temperature, including $t=0$, with threshold $34\,^{\circ}$C. No mechanical object geometry beyond this thermal face is assumed.

Native uses backward Euler at $N=160$ and $320$ with the fixed physical times above; the independent reference uses linear finite volumes and spectral exponential propagation at $N=640$. The reference peak changes by $0.00229\,^{\circ}$C from $N=320$ to $640$; native changes by $0.02367\,^{\circ}$C from $160$ to $320$. These are numerical checks, not physical uncertainty. The surface-matched uniform state for each history is
\begin{equation}
T_U=S+\frac{dz}{2k}\,h_8\,(S-25),
\end{equation}
where $S$ is the native natural-boundary reconstructed surface and $dz$ is the native cell thickness; all layers are set to $T_U$. This uses the native discrete half-cell reconstruction, preserves geometry, capacity and pad, removes only the depth state, is not energy-matched, and is not a reachable history. It is not fitted to the future.

The predeclared Cartesian grid is target surface $\{31,32\}\,^{\circ}$C $\times$ release $\{0.05,0.1,0.2,0.4\}$~s $\times$ pad capacity $\{0.5,1.0\}$~J/K. For each target/release pair, quench durations are matched using the N640 reference and then held identical for the native runs. A history is infeasible when its surface after zero quench and the specified release is already below the target; all four infeasible rows at release $0.4$~s are retained. Across the 12 feasible conditions (24 condition--history evaluations), full-layer native has peak MAE/max error $0.003454/0.024406\,^{\circ}$C and $0/24$ safety errors; the surface-matched uniform state has $4.340156/12.128856\,^{\circ}$C and $12/24$ safety errors. This deterministic sensitivity grid is not 24 independent physical samples, does not evaluate a learned policy, and does not establish hardware accuracy or generalization to unseen mechanisms.

\begin{table}[H]
\trrtablelayout
\centering\small
\caption{Hidden-depth sensitivity across 16 fixed combinations of target surface temperature, release duration and pad heat capacity. Cold and warm denote the two internal-storage histories. The four rows marked unreachable cannot attain the target surface temperature after release; the remaining rows compare future contact peaks.}
\label{tab:hidden-depth-grid-sup}
\setlength{\tabcolsep}{6pt}
\begin{tabular}{@{}rrrrrrrrr@{}}
\toprule
 & & & \multicolumn{6}{c}{\bfseries Future contact peak ($^{\circ}$C)} \\
\cmidrule(l){4-9}
 & & & \multicolumn{2}{c}{\bfseries Reference} & \multicolumn{2}{c}{\bfseries Full layers} & \multicolumn{2}{c}{\bfseries Uniform depth} \\
\cmidrule(lr){4-5}\cmidrule(lr){6-7}\cmidrule(l){8-9}
{\bfseries \shortstack{Target\\($^{\circ}$C)}} & {\bfseries \shortstack{Release\\(s)}} & {\bfseries \shortstack{Pad capacity\\(J/K)}} & {\bfseries Cold} & {\bfseries Warm} & {\bfseries Cold} & {\bfseries Warm} & {\bfseries Cold} & {\bfseries Warm} \\
\midrule
31 & 0.05 & 0.5 & 30.984 & 42.073 & 30.990 & 42.073 & 30.990 & 30.978 \\
31 & 0.05 & 1.0 & 30.984 & 42.033 & 30.990 & 42.033 & 30.990 & 30.978 \\
31 & 0.10 & 0.5 & 30.984 & 39.161 & 30.986 & 39.161 & 30.986 & 30.973 \\
31 & 0.10 & 1.0 & 30.984 & 39.127 & 30.986 & 39.128 & 30.986 & 30.973 \\
31 & 0.20 & 0.5 & 30.984 & 36.479 & 30.981 & 36.479 & 30.981 & 30.971 \\
31 & 0.20 & 1.0 & 30.984 & 36.452 & 30.981 & 36.452 & 30.981 & 30.971 \\
31 & 0.40 & 0.5 & \multicolumn{6}{c}{Target unreachable after release} \\
31 & 0.40 & 1.0 & \multicolumn{6}{c}{Target unreachable after release} \\
\addlinespace[4pt]
32 & 0.05 & 0.5 & 31.981 & 44.103 & 31.987 & 44.103 & 31.987 & 31.974 \\
32 & 0.05 & 1.0 & 31.981 & 44.057 & 31.987 & 44.058 & 31.987 & 31.974 \\
32 & 0.10 & 0.5 & 31.981 & 41.036 & 31.982 & 41.036 & 31.982 & 31.969 \\
32 & 0.10 & 1.0 & 31.981 & 40.998 & 31.982 & 40.998 & 31.982 & 31.969 \\
32 & 0.20 & 0.5 & 31.981 & 38.128 & 31.957 & 38.128 & 31.957 & 31.966 \\
32 & 0.20 & 1.0 & 31.981 & 38.096 & 31.957 & 38.097 & 31.957 & 31.966 \\
32 & 0.40 & 0.5 & \multicolumn{6}{c}{Target unreachable after release} \\
32 & 0.40 & 1.0 & \multicolumn{6}{c}{Target unreachable after release} \\
\bottomrule
\end{tabular}
\end{table}

Time-step refinement. We halved the short-heat, quench, release, and future steps at N320 (short heat $5\times10^{-5}$~s, quench/release $5\times10^{-5}$~s, future $2.5\times10^{-4}$~s); the 180~s heating phase retained its $5\times10^{-3}$~s step. The two baseline histories changed decision surface by at most $1.4\times10^{-3}\,^{\circ}$C and full peaks by at most $1.4\times10^{-3}\,^{\circ}$C. At the largest-error grid cell (target 32\,$^{\circ}$C, release 0.2~s), full peaks changed by less than $2.2\times10^{-4}\,^{\circ}$C. These checks quantify sensitivity to the refined integration steps.

\FloatBarrier
\section{Additional Mechanism Records}
\label{app:additional-mechanism-records}
These records detail the continuous transfer, surface histories and local placement decisions used to evaluate retained thermal state.
\subsection{Continuous execution and local decision details}
\label{app:execution-records-codex041}
In the continuous transfer, lifting exposes the source and destination surfaces at 174.8 and 217.3~s. Support and gripper contacts total 834.15~s within a 519.8~s execution because intervals overlap. The shell, supports and gripper histories are shown below. Contact labels use the native event clock; the exposed camera trace becomes available only after lifting.

\begin{figure}[H]
\centering
\includegraphics[width=0.65\linewidth]{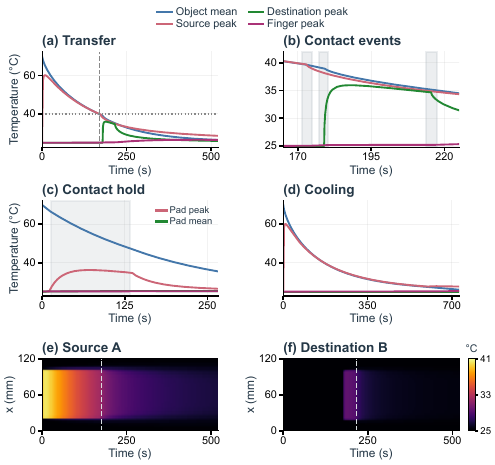}
\caption{Continuous thermal interaction record. (a--d) Transfer, contact detail, contact hold and cooling. (e,f) Source and destination surface histories averaged along $y$. Shading marks interaction intervals; dashed lines mark the trigger and uncovering.}
\label{fig:interaction-curves}
\end{figure}
\begin{table}[H]
\trrtablelayout
\centering\small
\caption{Surface-matched uniform state versus the full layer state on identical finite-pad recontact. Safe uses the $34\,^{\circ}$C peak rule. The uniform state is fitted to the current surface and is not energy-matched.}
\label{tab:hidden-depth-034}
\setlength{\tabcolsep}{2.5pt}
\begin{tabular}{lrrcc}
\toprule
 & \multicolumn{2}{c}{\bfseries Contact peak ($^{\circ}$C)} & \multicolumn{2}{c}{\bfseries Safe contact?} \\
\cmidrule(lr){2-3}\cmidrule(l){4-5}
{\bfseries Model} & {\bfseries Cold history} & {\bfseries Warm history} & {\bfseries Cold history} & {\bfseries Warm history} \\
\midrule
Independent reference & 31.905 & 40.976 & yes & no \\
\rowcolor{trrmethod}Full-layer native & 31.905 & 40.976 & yes & no \\
Uniform-depth native & 31.905 & 31.936 & yes & yes \\
\bottomrule
\end{tabular}
\end{table}
\begin{table}[H]
\trrtablelayout
\centering\small
\caption{Placement decisions under a common 33\,$^{\circ}$C payload-peak rule. The coupled scene supplies the outcome; the query uses evolving contact exchange or contact-off.}
\label{tab:interaction-decision}
\setlength{\tabcolsep}{3pt}
\begin{tabular}{@{}rrrllrll@{}}
\toprule
 & & \multicolumn{3}{c}{\bfseries Contact exchange disabled} & \multicolumn{3}{c}{\bfseries TRR coupled query} \\
\cmidrule(lr){3-5}\cmidrule(l){6-8}
{\bfseries \shortstack{Vessel\\($^{\circ}$C)}} & {\bfseries \shortstack{Peak if placed\\($^{\circ}$C)}} & {\bfseries \shortstack{Predicted\\peak ($^{\circ}$C)}} & {\bfseries Action} & {\bfseries Outcome} & {\bfseries \shortstack{Predicted\\peak ($^{\circ}$C)}} & {\bfseries Action} & {\bfseries Outcome} \\
\midrule
40\,$^{\circ}$C & 27.57 & 25.00 & Place & Safe & 27.57 & Place & Safe \\
50\,$^{\circ}$C & 29.26 & 25.00 & Place & Safe & 29.26 & Place & Safe \\
60\,$^{\circ}$C & 30.93 & 25.00 & Place & Safe & 30.93 & Place & Safe \\
70\,$^{\circ}$C & 32.58 & 25.00 & Place & Safe & 32.58 & Place & Safe \\
80\,$^{\circ}$C & 34.22 & 25.00 & Place & Too hot & 34.22 & Avoid & Safe \\
90\,$^{\circ}$C & 35.85 & 25.00 & Place & Too hot & 35.85 & Avoid & Safe \\
\bottomrule
\end{tabular}
\end{table}
The placement sweep uses a 150~J/K payload initially at $25\,^{\circ}$C and holds each vessel for 60~s. The largest safe and smallest violating peaks are $32.58/34.22\,^{\circ}$C. At thresholds of 30 and $36\,^{\circ}$C, contact-off makes four and zero errors, respectively, while the coupled query makes no errors on these six coupled outcomes. The runs share 414 contact records and a common camera schedule; the coupled relative energy residual is below $10^{-11}$. These are consequences of the simulated state, not independent physical measurements.
\section{Reproducibility Manifest}
\label{app:release}
The release package is organized into six parts: (i) simulator core with layered assets, contact exchange, state transport and the camera/observation model; (ii) asset and measured-calibration data; (iii) task definitions and label generation; (iv) mixed-corpus manifests and fixed train/test splits; (v) training and evaluation scripts for forecasting and policy diagnostics; and (vi) scripts that regenerate each table and figure. Each study records its source split, training seed or deterministic model specification, evaluation cohort, and code/model versions. Saved per-decision records retain the inputs, choices and execution references used to regenerate the reported results. Environment versions and command lines are listed in the release README. Together, the saved records and regeneration scripts identify the configuration, seed and scoring unit behind each reported table and figure.

\nocite{li2023behavior,walke2023bridgedata,mees2022calvin,brahmbhatt2019contactdb,brahmbhatt2020contactpose,lin2025constraint,xue2025demogen,jiang2025dexmimicgen,si2024difftactile,dirsigthermal,brown2010verification,khazatsky2024droid,wang2026etgs,wagner2026fire,gazebothermal,savva2019habitat,zhou2024hazard,su2026hearth,hu2021radiation,li2021igibson,makoviychuk2021isaac,hwang2015multispectral,liu2023libero,jia2021llvip,gu2023maniskill2,tao2025maniskill3,ha2017mfnet,mandlekar2023mimicgen,zhou2025motif,todorov2012mujoco,kim2025well,oktal2022,o2024open,gao2026physir,tang2022piafusion,shivakumar2020pst900,fu2022rfuniverse,9001253,nasiriany2024robocasa,zhu2020robosuite,chen2025robotwin,yu2026safe,xiang2020sapien,rs4082477,son2026selective,grimaldi2502stonefish,chung2026tactile,wang2022tacto,bhattacharjee2021material,kaufmann2023expanding,bhattacharjee2015material,lin2024thermalnerf,ye2024thermal,zhang2023interactive,son2026thermoact,hassan2024thermonerf,pedro2021thrend}

\label{LastPage}
\end{document}